\documentclass[10pt,twocolumn,letterpaper]{article}

\usepackage{iccv}
\usepackage{times}
\usepackage{epsfig}
\usepackage{graphicx}
\usepackage{amsmath}
\usepackage{amssymb}

\usepackage{color}
\usepackage{tabularray}

\usepackage{glossaries}
\usepackage{dirtytalk}
\usepackage{caption}
\usepackage{subcaption}
\usepackage{setspace}
\usepackage{graphicx}

\usepackage{algorithm}
\usepackage[noend]{algpseudocode}

\usepackage[export]{adjustbox}

\usepackage[pagebackref=true,breaklinks=true,letterpaper=true,colorlinks,bookmarks=false]{hyperref}

 \iccvfinalcopy 
\def\iccvPaperID{2} 

\ificcvfinal\fi

\newcommand\footnoteref[1]{\protected@xdef\@thefnmark{\ref{#1}}\@footnotemark}

\newcommand{\R}{\mathbb{R}}

\newcommand{\minnumimg}{N_{\text{Exp3}}}

\newcommand{\numdim}{d}
\newcommand{\numdimred}{k}

\newcommand{\features}[1]{\texttt{features.{#1}}}

\newcommand{\img}{I}
\newcommand{\imgnumdim}{n}

\newcommand{\imgset}{\mathbf{I}}
\newcommand{\imgtrain}{\imgset_{\text{train}}}

\newcommand{\fvec}{x}
\newcommand{\fvecset}{\mathbf{X}}
\newcommand{\fvectrain}{\fvecset_{\text{train}}}

\newcommand{\cnn}{f}
\newcommand{\node}{N}
\newcommand{\fx}{\cnn_{\node}}

\newcommand{\mean}{\mu}
\newcommand{\empmean}{\hat{\mean}}

\newcommand{\covmat}{\Sigma}
\newcommand{\empcovmat}{\hat{\covmat}}

\newcommand{\precmat}{\covmat^{-1}}
\newcommand{\empprecmat}{\empcovmat^{-1}}

\newcommand{\gdistrib}{\mathcal{N}}

\newcommand{\md}{D_M}

\newcommand{\evec}{q}
\newcommand{\evecmat}{Q}
\newcommand{\evecset}{\mathbf{Q}}
\newcommand{\evalmat}{\Lambda}
\newcommand{\gfunc}{g}

\newcommand{\wvec}{w}
\newcommand{\wvecset}{\mathbf{W}}
\newcommand{\wvectest}{\wvecset_{\text{test}}}
\newcommand{\wvecgreedy}{\wvecset_{\text{greedy}}}
\newcommand{\wveceval}{\wvecset_{\text{eval}}}

\newcommand{\euclideanNorm}[1]{\lVert \, #1 \, \rVert_2}
 
\newcommand{\evecin}{\evecset^{\text{in}}}
\newcommand{\evecout}{\evecset^{\text{out}}}

\DeclareMathOperator*{\argmax}{argmax} 

\newglossaryentry{mvg}{
    name={MVG},
    first={multivariate Gaussian (MVG)},
    description={}
}

\newglossaryentry{pca}{
    name={PCA},
    first={Principal Component Analysis (PCA)},
    description={}
}

\newglossaryentry{npca}{
    name={NPCA},
    first={\textit{Negated} Principal Component Analysis (NPCA)},
    description={}
}

\newglossaryentry{cnn}{
    name={CNN},
    first={convolutional neural network (CNN)},
    description={}
}

\newglossaryentry{md}{
    name={M. distance},
    first={Mahalanobis distance (M. distance)},
    description={}
}

\newglossaryentry{ad}{
    name={AD},
    first={Anomaly detection (AD)},
    description={}
}

\newglossaryentry{auroc}{
    name={AUROC},
    first={Area Under the Receiver Operating Characteristic curve (AUROC)},
    description={}
}

\newglossaryentry{ledoitwolf}{
    name={LeDoit-Wolf},
    description={}
}

\newglossaryentry{mvtecad}{
    name={MVTec-AD},
    first={MVTec Anomaly Detection (MVTec-AD)},
    description={},
}

\newglossaryentry{imgnet}{
    name={ImageNet},
    description={},
}

\newglossaryentry{bottomup}{
    name={Bottom-Up},
    description={},
}

\newglossaryentry{topdown}{
    name={Top-Down},
    description={},
}

\newglossaryentry{effnet}{
    name={EfficientNet},
    description={},
}

\newglossaryentry{effnetb0}{
    name={EfficientNet B0},
    description={},
}

\newglossaryentry{gaussianad}{
    name={Gaussian-AD},
    description={},
}

\begin{document}


\title{Gaussian Image Anomaly Detection with Greedy Eigencomponent Selection}

\author{Tetiana Gula \\
{Centre for mathematical morphology}\\
{{Mines Paris - PSL}}\\
Fontainebleau, France \\
{\tt\small tetiana.gula@etu.minesparis.psl.eu}
\and
João P. C. Bertoldo \\
{Centre for mathematical morphology}\\
{{Mines Paris - PSL}}\\
Fontainebleau, France \\
{\tt\small jpcbertoldo@minesparis.psl.eu}
}


\maketitle

\ificcvfinal\thispagestyle{empty}\fi


\begin{abstract}
    \gls{ad} in images, identifying significant deviations from normality, is a critical issue in computer vision. This paper introduces a novel approach to dimensionality reduction for \gls{ad} using pre-trained \gls{cnn} that incorporate \gls{effnet} models. We investigate the importance of component selection and propose two types of tree search approaches, both employing a greedy strategy, for optimal eigencomponent selection. Our study conducts three main experiments to evaluate the effectiveness of our approach. The first experiment explores the influence of test set performance on component choice, the second experiment examines the performance when we train on one anomaly type and evaluate on all other types, and the third experiment investigates the impact of using a minimum number of images for training and selecting them based on anomaly types. Our approach aims to find the optimal subset of components that deliver the highest performance score, instead of focusing solely on the proportion of variance explained by each component and also understand the components behaviour in different settings. Our results indicate that the proposed method surpasses both \gls{pca} and \gls{npca} in terms of detection accuracy, even when using fewer components. Thus, our approach provides a promising alternative to conventional dimensionality reduction techniques in \gls{ad}, and holds potential to enhance the efficiency and effectiveness of \gls{ad} systems.
\end{abstract}


\glsresetall


\section{Introduction}
\gls{ad} is a challenging task in machine learning with a wide range of applications, from fraud detection in financial transactions to fault diagnosis in industrial systems. 
In recent years, deep learning approaches have shown promising results in detecting anomalies from images, particularly using pre-trained \gls{cnn}. 
However, one of the key challenges is that \gls{cnn} can produce a large number of features, which can lead to computational challenges, and the presence of redundant information may not contribute the detection task. 

In this work, our focus is on dimensionality reduction.
Previous studies have utilized dimensionality reduction techniques such as \gls{pca}, which select a subset of components that capture the most variation in the data, its variant \gls{npca} introduced by \cite{rippel_gaussian_2021}, random feature selection \cite{defard_padim_2021}, and random linear projections \cite{kim_semi-orthogonal_2021}.

We experiment with several strategies using a \gls{mvg} model trained on image features extracted from a pre-trained \gls{cnn}, as proposed in \cite{rippel_gaussian_2021}, on the well-established \gls{mvtecad} dataset \cite{bergmann_mvtec_2019, bergmann_mvtec_2021}, which focuses on inspection tasks, presenting challenging real-world use-cases for \gls{ad}.

Specifically, our investigation focuses on the potential of using eigendecomposition combined with component selection, for which we use a greedy tree search approach.
We introduce two types of tree traversal modes, namely Bottom-Up and Top-Down. 
The Bottom-Up approach gradually adds components that yield the best performance, while the Top-Down approach gradually removes components that do not contribute to the performance.

Our strategy aims to find the optimal subset of components that delivers the best performance score, rather than focusing solely on the proportion of variance captured by each component -- the premise used in \gls{pca} and \gls{npca}.


We test our algorithm on an ideally supervised setting revealing that it is \textit{possible} to significantly outperform both \gls{pca} and \gls{npca} with a much smaller embedding space.
The two types of tree traversal modes - Bottom-Up and Top-Down - remarkably align and reveal a substantial redundancy in the subspaces identified as contributory, which underscores the importance of dimensionality reduction.
In contrast, our experiments show that generalization is not easily achievable even scenarios where the performance improvement is seemingly easy. 

Overall, this highlights that our proposed approach provides a promising alternative to traditional dimensionality reduction techniques in the field of \gls{ad}.

\section{Methods}\label{sec:methods}

We use the \gls{mvg}-based model proposed by \cite{rippel_modeling_2021}, but, while it originally uses \gls{pca} or \gls{npca} to reduce the number of dimensions, we propose a more general technique, likewise based on the eigendecomposition of the covariance matrix.

We propose to approximate the optimal subspace of eigencomponents without imposing any restriction to it and directly optimizing a performance function -- like a feature selection algorithm.
We further introduce an intermediate whitening operation to simplify the eigencomponent selection.

\subsection{Multivariate Gaussian (MVG)}\label{sec:mvg}

A data point $\img \in \R^\imgnumdim$ (with size $\imgnumdim$) is passed as an input to a neural network $\cnn$, referred to as \say{backbone}; 
then, the internal activations of its node\footnote{The term \say{node}, instead of \say{layer}, is used to match \texttt{pytorch}.} $\node$ are extracted.
This operator is further referred the \say{feature extractor} $\fx: \R^\imgnumdim \to \R^\numdim$, where $\fvec = \fx(\img)$ is the feature vector of $\img$. 

Assuming that the feature vectors extracted from a set of \textit{normal} data points $\imgtrain$ follow an \gls{mvg} distribution $\gdistrib(\mean, \covmat)$, with mean $\mean$ and covariance matrix $\covmat$, anomalous data points are likely to lie far away from the mean of this distribution, where the notion of distance is measured with the \gls{md}:

\begin{equation}\label{eq:md}
    \md(\fvec) = \sqrt{(\fvec - \mean)^T \precmat (\fvec - \mean)}
\end{equation}

Using the set of normal feature vectors $\fvectrain$ (extracted from $\imgtrain$), the empirical mean vector $\empmean \in \R^\numdim$ is fitted with the maximum likelihood estimator $\empmean = \frac{1}{| \fvectrain |} \sum_{\fvec \in \fvectrain} \fvec$, and the empirical covariance matrix $\empcovmat \in \R^{\numdim \times \numdim}$ is fitted using \gls{ledoitwolf}'s method.
This estimator ensures a positive definite inverse covariance matrix $\empprecmat$ by adding a regularization term to the maximum likelihood estimator while automatically selecting the optimal regularization parameter based on the number of observations and features in the dataset, achieving a balance between bias and variance.


\subsection{Whitening}\label{sec:whitenning}

As $\empcovmat$ is a real symmetric matrix, it can be decomposed as $\empcovmat = \evecmat \evalmat \evecmat^T$, where $\evecmat$ is an orthogonal matrix with column $\evec_i$ being the $i$-th eigenvector of $\empcovmat$, and $\evalmat$ is the diagonal matrix with the element $\evalmat_{ii} = \lambda_{i}$ being the $i$-th eigenvalue of $\empcovmat$.
Since the regularization of $\empcovmat$ ensures that its eigenvalues are real and positive, the whitening matrix $\evalmat^{-\frac{1}{2}} \evecmat^T$, where the inverse square root is taken elementwise (because $\evalmat$ is diagonal), can be used to build white feature vectors 

\begin{equation}\label{eq:wvec}
    \wvec = \left( \evalmat^{-\frac{1}{2}} \evecmat^T \right) \left( \fvec - \empmean \right) \quad ,
\end{equation}

\noindent
such that the linear projection on the left is $(\empprecmat)^{\frac{1}{2}}$, the square root matrix (\textit{not} elementwise) of the empirical precision matrix (inverse of the empirical covariance matrix): 

\begin{equation}\label{eq:prec-mat-squared-matrix}
\begin{split}
    \left( \evalmat^{-\frac{1}{2}} \evecmat^T \right)^2 
    & = \left( \evalmat^{-\frac{1}{2}} \evecmat^T \right) \left( \evalmat^{-\frac{1}{2}} \evecmat^T \right) \\
    & = \evecmat \left( \evalmat^{-\frac{1}{2}} \evalmat^{-\frac{1}{2}} \right) \evecmat^T \\
    & = \evecmat \evalmat^{-1} \evecmat^T 
      = \empprecmat 
\end{split}
\end{equation}

\begin{equation}\label{eq:prec-mat-squared-matrix-therefore}
    \therefore \left( \evalmat^{-\frac{1}{2}} \evecmat^T \right) = (\empprecmat)^{\frac{1}{2}}  
    \quad .
\end{equation}

\noindent
Therefore the \gls{md} (Equation~\ref{eq:md}) can be expressed as $\euclideanNorm{\wvec}$:

\begin{equation}\label{eq:md-from-wvec}
\begin{split}
    \md(\fvec)^2
     & = (\fvec - \empmean)^T \left[ (\empprecmat)^{\frac{1}{2}} (\empprecmat)^{\frac{1}{2}} \right] (\fvec - \empmean) \\
     & = \left[ (\empprecmat)^{\frac{1}{2}} (\fvec - \empmean) \right]^T \left[ (\empprecmat)^{\frac{1}{2}} (\fvec - \empmean) \right] \\
     & = \wvec^T \wvec = \euclideanNorm{\wvec}^2 \\
\end{split}
\end{equation}

\begin{equation}\label{eq:md-from-wvec-therefore}
     \therefore \md(\fvec) = \euclideanNorm{\wvec}
     \quad ,
\end{equation}

\noindent
where $\euclideanNorm{\cdot}$ is the Eucledian norm.
In short, Equation~\ref{eq:wvec} provides an alternative to compute the anomaly score $\md(\fx(\img))$ as $\euclideanNorm{ \evalmat^{-\frac{1}{2}} \evecmat^T \left( \fx(\img) - \empmean \right)}$.

Notice that the entry $\wvec_{i}$ (the $i$-th row of the white vector $\wvec$) corresponds to the projection of the centered feature vector $(\fvec - \empmean)$ onto the eigenvector $\evec_i$.  
The axes of the vector space of $\wvec$ are further referred to as \textit{components} to remind that they come from the eigendecomposition of $\empcovmat$.


\subsection{Greedy Eigencomponent Selection}\label{sec:dimred}


Let $\evecset = \{ \evec_1 , \dots , \evec_\numdim \}$ be the set of eigenvectors from $\empcovmat$, $\evecset_\numdimred \subseteq \evecset$ a subset of eigenvectors such that $ | \evecset_\numdimred | = \numdimred $, and $\gfunc$ a performance function (higher is better) that evaluates the quality of a dimension reduction choice.
Our proposed framework consists of finding the optimal subset of eigenvectors (eigenvalues are omitted for the sake of simplicity)

\begin{equation}
    \evecset_\numdimred^{*} = \argmax_{\evecset_\numdimred \subseteq \evecset } \; \gfunc( \evecset_\numdimred ) \quad .
\end{equation}

However, finding the optimal subset of $\numdimred$ components $\evecset_\numdimred^{*}$ is a combinatorial problem with a search space of size $\numdim$-choose-$\numdimred$.
To make this problem amenable, we propose to approximate this optimization with a \textit{greedy} algorithm, which consists of iteratively building the optimal subset one component at a time while locally optimizing $\gfunc$.

The greedy eigencomponent selection can be carried out in two ways: starting with an empty set then adding components, which we call the \say{bottom-up} variant (a.k.a. forward variable selection), or starting with the set of all components then removing components, which we call the \say{top-down} variant (a.k.a. backward variable selection).
Algorithm~\ref{alg:greedy-bottom-up} describes the bottom-up variant and Algorithm~\ref{alg:greedy-top-down} describes the top-down variant.

\begin{algorithm}[!t]
\caption{Greedy Bottom Up}\label{alg:greedy-bottom-up}
\begin{algorithmic}[1]
\Require $\numdim = | \evecset | > 0$
\Require $1 \leq \numdimred \leq \numdim $
\Procedure{GreedyBottomUp($\evecset$, $\numdimred$, $\gfunc$)}{}
    \State $\evecin \gets \emptyset \quad \; \; \, \text{  \# set of eigenvectors \textit{IN} the model}$  
    \State $\evecout \gets \evecset \quad \text{  \# set of eigenvectors \textit{OUT} of the model}$  
    \While{$ |\evecin| < \numdimred$}
        \State $\evec^* \gets \argmax_{\evec \in \evecout} \; \gfunc\left( \evecin \, \cup \, \{ \evec^* \} \right)$
        \State $\evecin \gets \evecin \, \cup \, \{ \evec^* \}$
        \State $\evecout \gets \evecout \setminus \{ \evec^* \}$
    \EndWhile
\EndProcedure
\end{algorithmic}
\end{algorithm}

\begin{algorithm}[!t]
\caption{Greedy Top Down}\label{alg:greedy-top-down}
\begin{algorithmic}[1]
\Require $\numdim = | \evecset | > 0$
\Require $1 \leq \numdimred \leq \numdim $
\Procedure{GreedyTopDown($\evecset$, $\numdimred$, $\gfunc$)}{}
    \State $\evecin \gets \evecset \quad \text{  \# set of eigenvectors \textit{IN} the model}$  
    \State $\evecout \gets \emptyset \quad \text{  \# set of eigenvectors \textit{OUT} of the model}$  
    \While{$ |\evecin| > \numdimred$}
        \State $\evec^* \gets \argmax_{\evec \in \evecin} \; \gfunc\left( \evecin \setminus \{ \evec^* \} \right)$
        \State $\evecin \gets \evecin \setminus \{ \evec^* \}$
        \State $\evecout \gets \evecout \, \cup \, \{ \evec^* \}$
    \EndWhile
\EndProcedure
\end{algorithmic}
\end{algorithm}

\begin{figure}[!t]
    \centering
    \includegraphics[width=0.5\textwidth]{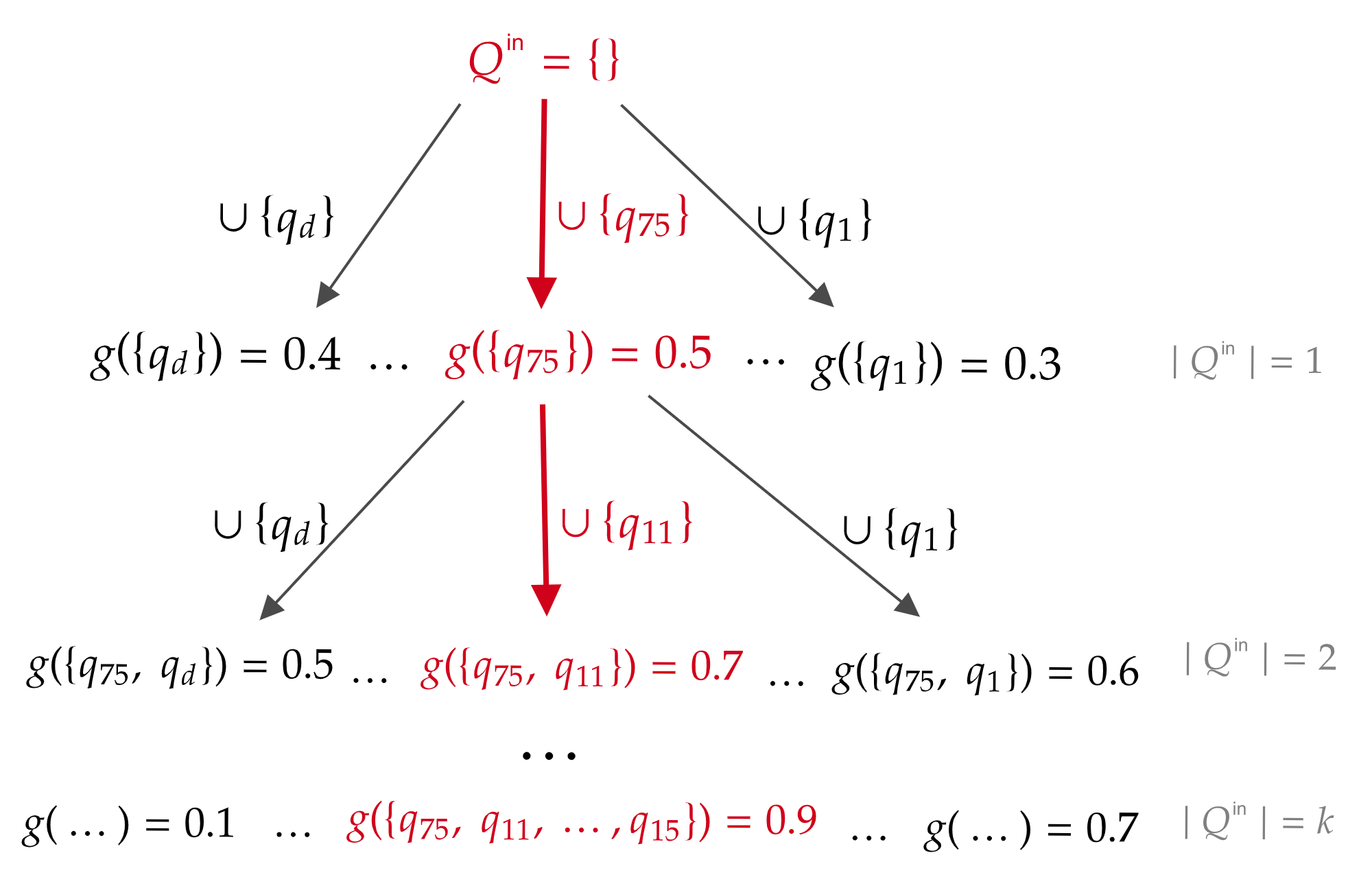}
    \caption{Greedy Bottom Up algorithm illustrated as tree search. The red path represent the chosen path at each step optimization.}
    \label{fig:greedy-bottom-up}
\end{figure}

In simple terms, the bottom-up approach (illustrated with a search tree representation in Figure~\ref{fig:greedy-bottom-up}) starts by selecting the component that yields the best performance score (given by the function $\gfunc$).
Then, we search for the second component that, when combined with the first one, yields the best performance score.
This process is repeated until the predetermined number of components $\numdimred$ has been reached.

In practice, the whitening transformation (Equation~\ref{eq:wvec}) makes it simple to obtain a reduced model with $\evecset^\prime \subseteq \evecset$.
Assuming eigenvalues (and respective eigenvectors) to be sorted in increasing order, the entries of $\wvec$ are equally sorted because the rows of $\left( \evalmat^{-\frac{1}{2}} \evecmat^T \right)$ follow the same order.
Therefore a reduced model can be reproduced by choosing the entries from $\wvec$ with the same indexes of the selected eigencomponents.
Example: the component selection $\evecset^\prime = \{ \evec_{75} \, , \, \evec_{11} \, , \, \evec_{15} \}$ corresponds to using, as anomaly score,  $\euclideanNorm{\wvec^\prime} = \euclideanNorm{ \left( \wvec_{75} \, , \, \wvec_{11} \, , \, \wvec_{15} \right)^T }$

\paragraph{Previous works}
\gls{npca} with $\numdimred$ output dimensions corresponds to $\wvec_{1:\numdimred} = \left( \wvec_1, \dots, \wvec_{\numdimred} \right)^T$, which is the truncation of $\wvec$ to its first $\numdimred$ entries -- \gls{pca} analogously corresponds to truncating $\wvec$ to its \textit{last} $\numdimred$ entries.
The subspace decomposition proposed in \cite{lin_deep_2022} corresponds to splitting $\wvec$ in three: $\left[ w_{1:m_2} , w_{m_2:m_1}, w_{m_1:\numdim} \right]$ -- where $1 \leq m_1 \leq m_2 \leq d$ are two breakpoints heuristically defined.

\section{Experimental setup}\label{sec:experimental-setup}


We establish a general setup to study our greedy eigencomponent selection algorithm then experiment with three scenarios using different data splits for the algorithm execution and evaluation.

\subsection{Dataset}\label{sec:dataset}

We use the \gls{mvtecad} dataset \cite{bergmann_mvtec_2019, bergmann_mvtec_2021}\footnote{We deliberately do not show image samples for the sake of space, which can easily be found at \href{https://www.mvtec.com/company/research/datasets/mvtec-ad}{\nolinkurl{www.mvtec.com/company/research/datasets/mvtec-ad}}.}. 
It comprises 15 categories with 3629 normal images for training and 1725 images for testing. 

Each category is used independently (as if they were 15 independent datasets), with its training set containing only defectless (normal) images , while its test set contains both normal and anomalous images, which are from a variety of defects, such as surface scratches, dents, distorted or missing object parts, etc. 
The defects were manually generated to produce realistic anomalies as they would occur in real-world industrial inspection scenarios.
In total, 73 different defect types, or \say{anomaly types}, are present.
For more details, refer to Table~\ref{tab:num_images} in the Appendix~\ref{sec:mvtec-num-images}.

\subsection{Feature extractor}\label{sec:fx}

Like \cite{lin_deep_2022}, we choose the \gls{effnet} family of models as backbone for our feature extractor.
Specifically, we use \gls{effnetb0} with the pre-trained weights \texttt{EfficientNet\_B0\_Weights.IMAGENET1K\_V1} from \texttt{torchvision}, which were pre-trained on classification on \gls{imgnet}.

We analyze the nine main nodes from \gls{effnetb0}, which are sequentially named from \say{\features{0}} to \say{\features{8}}\footnote{Other authors have named them from 1 to 9.}.
As the internal activations of a \gls{cnn} come as a 3D tensor (channels, width, and height axes), we apply a global average pooling on the spatial axes (along the width and height) to obtain an image-wise feature vector.

\subsection{Metrics}\label{sec:metrics}

We use the \gls{auroc} both to optimize the eigencomponent selection (i.e. the $\gfunc$ function) and to measure the models' performances.
The \gls{auroc} score is a widely used metric for anomaly detection and conveniently threshold selection-free, unlike binary classification metrics like the accuracy. 
It measures the ability of a model to discriminate between normal and anomalous instances, with a score of 0.5 indicating random guessing and a score of 1.0 indicating perfect discrimination.

\subsection{Data split}\label{sec:data-split}

The train set is used to fit the \gls{mvg} model $\gdistrib(\empmean, \empcovmat)$, which is then used to generate $\wvectest$, the set of white vectors (Equation~\ref{eq:wvec}) from the test set.

While the train set is fully (and only) used at the first step, the test set is further split in two: $\wvecgreedy$ and $\wveceval$.
The \say{greedy white vectors} $\wvecgreedy$ are used in the function $\gfunc$ to guide the search tree traversal (Algorithms~\ref{alg:greedy-bottom-up} and~\ref{alg:greedy-top-down}), and the \say{evaluation white vectors} $\wveceval$ are used to measure the models' performances (the values reported in the results).

It's worth noting that we use the test set, not the training set, for both $\wvecgreedy$ and $\wveceval$ because they must contain normal \textit{and} anomalous instances so that \gls{auroc} can be computed.
In a fully unsupervised scenario, we wouldn't have access to anomalous samples -- although it could be simulated with synthetic anomalies.
However, we focus on using real anomalies for the sake of leveraging or framework to gain insights into the \gls{mvg} model.

\subsection{Other details}\label{sec:other-details}

Both greedy modes, botom-up and top-down, are evaluated in all the scenarios (category-node combinations).
However, some analyses may present only the results of one mode for brevity, which, by default, is bottom-up. 

Every scenario and dimension reduction strategy is evaluated using all the possible values of $\numdimred \in \{1 \, , \, \dots \, , \, \numdim \}$, where $\numdim$ ranges from a few 10s in the shallowest nodes, to 100s in the deeper nodes, up to $\sim$1000 in the last node.
Then, the results are plotted as curves of $\numdimred$ vs. \gls{auroc}, as in Figure~\ref{fig:exp1_selection}, Figure~\ref{fig:exp2_selection}, and Figure~\ref{fig:exp3_selection}.
The lines are plotted with the number of eigencomponents $\numdimred$ on the X-axis and the respective \gls{auroc} on $\wveceval$ on the Y-axis.

For the sake of space, the graphs from all scenarios are documented in the Appendix, while we show representative cases in the main text.

Section~\ref{sec:exp1-overfit}, Section~\ref{sec:exp2-peranomtype}, and Section~\ref{sec:exp3-fixednumimgs} present different data split configurations, which essentially change the meaning of the results, so their discussions are also presented separately. 

\section{Experiment 1: test set overfit}\label{sec:exp1-overfit}

\subsection{Experiment 1 Setup}

We set $\wvecgreedy = \wveceval = \wvectest$, which is an intentional overfit of the test set (both component optimization and evaluation use the full test data).
Despite this scenario being unrealistic, it is useful for diagnostic and research purposes.

By overfiting the test set with our algorithm, we can measure its potential and compare it with the results achieved by previous approaches (\gls{pca}, \gls{npca}, and \cite{lin_deep_2022}'s subspace decomposition).
Furthermore, as our results show, this setup reveals interesting insights into the task and into the feature extractor.

In Experiment 2 (Section~\ref{sec:exp2-peranomtype}) and Experiment 3 (Section~\ref{sec:exp3-fixednumimgs}), $\wvecgreedy$ and $\wveceval$ do \textit{not} have anomalous images in common so the generalization power of our proposed framework can be compared to the \textit{achievable} performances revealed by Experiment 1.
In particular, we focus Experiment 2 and Experiment 3 on the nodes from \features{5} up to \features{8} because they show better achievable performance -- often perfect (i.e. \gls{auroc} 100\%) -- and more stable behavior.

\subsection{Experiment 1 Results}

\begin{figure}[!h]
    \centering
    \begin{subfigure}[b]{\linewidth}
        \centering
        \adjincludegraphics[
            width=\textwidth,
            trim={{0.12\width} {0.860\height} {0.662\width} {0.011\height}},clip,
            keepaspectratio,
        ]{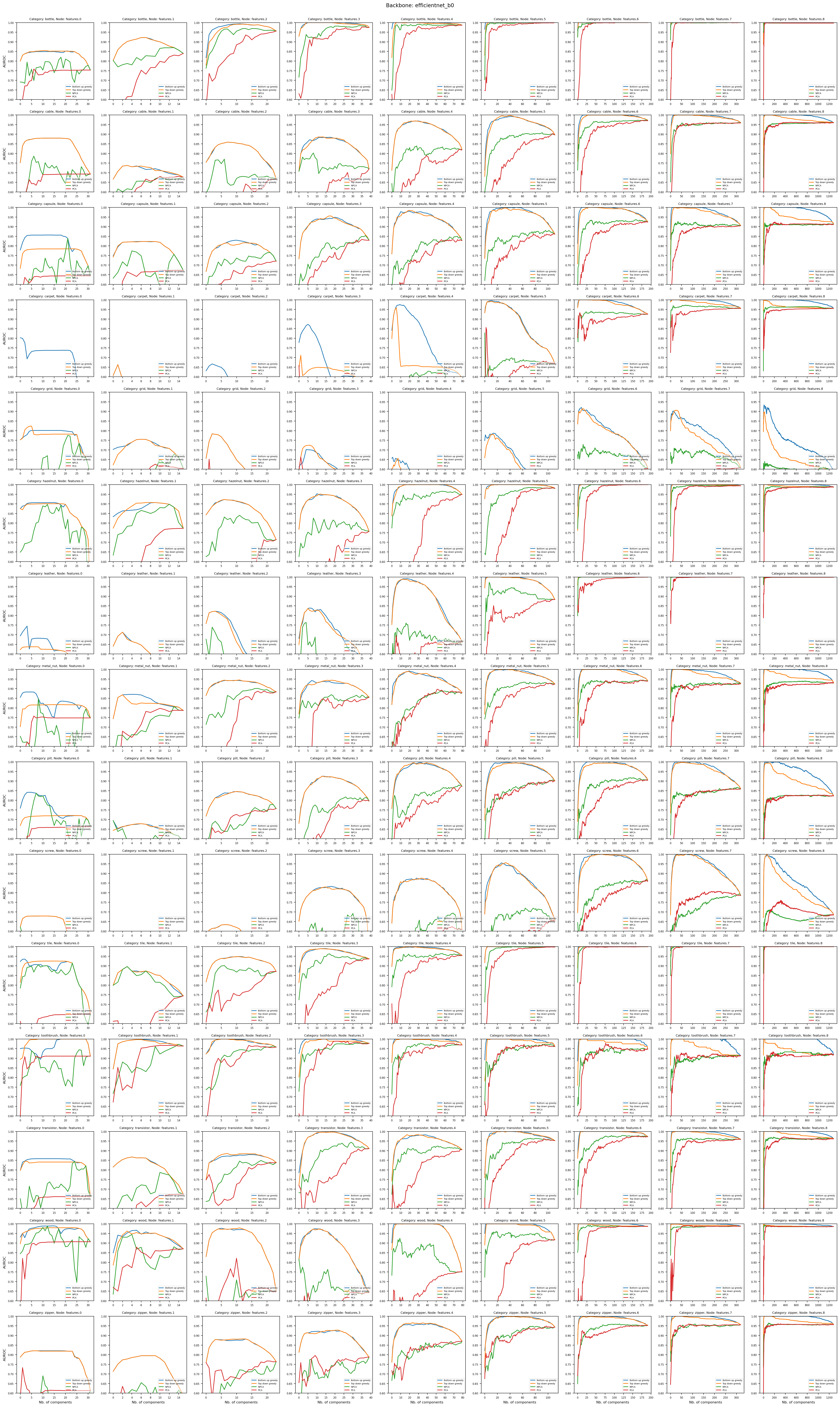}
        \caption{\{ bottle, cable \} $\times$ \{ \features{1}, \features{2} \}}
        \label{fig:exp1_selection__bottle_cable_f1_f2}
    \end{subfigure}
    \hfill
    \begin{subfigure}[b]{\linewidth}
        \centering
        \adjincludegraphics[
            width=\textwidth,
            trim={{0.67\width} {0.332\height} {0.11\width} {0.535\height}},clip,
            keepaspectratio,
        ]{figures/full_depth.png}
        \caption{\{ pill, screw \} $\times$ \{ \features{6}, \features{7} \}}
        \label{fig:exp1_selection__pill_screw_f6_f7}
    \end{subfigure}
    \caption{
        Experiment 1: selection of representative cases.
    }
    \label{fig:exp1_selection}
\end{figure}

\begin{figure*}[!t]
    \centering
    \includegraphics[width=0.9\textwidth]{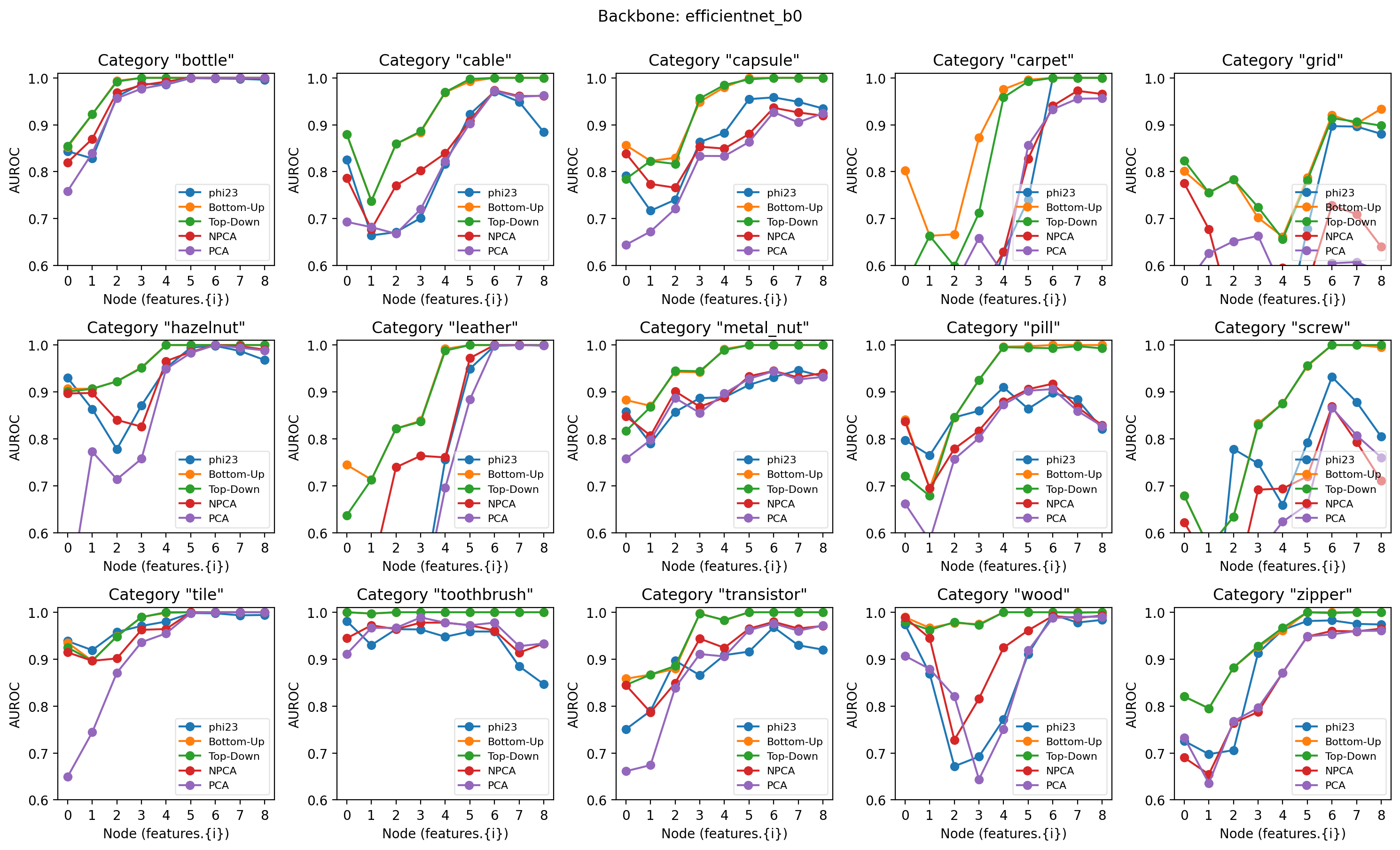}
    \caption{
        Experiment 1: best AUROC out of all values of $\numdimred$ per node.
        The curve "phi23" refers to the results from \cite{lin_deep_2022} with the alternative "$\left[ \Phi_{2}, \Phi_{3} \right]$".
    }
    \label{fig:exp1_best_auroc}
\end{figure*}

The results of Experiment 1 for all scenarios can be found in Appendix~\ref{sec:exp1_all_graphs}, and Figure~\ref{fig:exp1_selection} shows a selection of representative cases.
Both traversal modes, Bottom-Up and Top-Down, are compared with \gls{pca} and \gls{npca}. 
Notice that at the point where $\numdimred = \numdim$ (i.e. no dimension reduction) all the lines merge to the same point because they simplify to the same model.

Figure~\ref{fig:exp1_best_auroc} shows a summary plot with the node index on the X-axis and the best \gls{auroc} achieved by that node on the Y-axis -- each graph corresponds to picking a row (category) from Figure~\ref{fig:exp1_all_graphs} and extracting the maximum \gls{auroc} for each node curve.
We additionally compare the with the results achieved in \cite{lin_deep_2022}\footnote{We thank the authors for having shared their results data with us.}.
Since \cite{lin_deep_2022} proposes several subspace decompositions (i.e. several dimension reductions), we selected the alternative \say{$\left[ \Phi_{2}, \Phi_{3} \right]$} -- which is equivalent to \gls{npca} with a heuristic choice of $\numdimred$ -- because it achieves the best results in most scenarios.

\subsection{Experiment 1 Discussion}

Our analysis suggests that it is \textit{possible} to (sometimes greatly) enhance the performance of the \gls{mvg} model by cherry-picking eigencomponents from $\empcovmat$ -- Experiment 2 (Section~\ref{sec:exp2-peranomtype}) and Experiment 3 (Section~\ref{sec:exp3-fixednumimgs}), however, show that this same approach unfortunately fails to generalize well.
Furthermore, we find that the deeper layers require fewer components to achieve high-performance scores (see Appendix~\ref{sec:k_at_max_auroc} for more details).

Most scenarios using the \gls{bottomup} mode exhibit a gradual increase in performance until reaching a certain level of saturation, after which they exhibit diminishing performance.
This behavior is further analyzed in Appendix~\ref{sec:regimes}, where we split the $\numdimred$-vs-\gls{auroc} curves in these three respective regimes: rise, plateau, and drop.
Some scenarios, however, show an edge case behavior with an rise regime followed directly by a quick drop of performance (e.g. categories \say{metal nut}, \say{pill}, and \say{screw} with \features{8}, and category grid with the last four nodes).

Comparing \gls{pca} and \gls{npca} with the greedy eigencomponent selection, it's clear that the former -- even at their optimal point -- are generally much bellow the achievable performance demonstrated by latter (see Figure~\ref{fig:exp1_selection}).
As the results demonstrate, high performance can be achieved using only 30-40 components, and even nearly-perfect class discrimination (100\% \gls{auroc}) with less than 10 components (see categories \say{bottle}, \say{carpet}, \say{hazelnut}, \say{leather}, \say{toothbrush}, \say{tile}, \say{wood} in the Appendix~\ref{sec:regimes}). 

In other words, it is possible to achieve high performance in \gls{ad} with rather small embeddings. 
These findings underscore the importance of dimension reduction, particularly for deeper layers of the network, where the total number of components can be substantial (100s or even 1000s).

In contrast, \gls{pca} and \gls{npca} require a relatively large number of components to achieve their best score, which is systematically lower than that achieved using the Greedy \gls{auroc} method with 30$\sim$40 components.
We conjecture that the limited performance of \gls{pca} and \gls{npca} are attributed to the constraint imposed to select components based on their variance, which results in selecting irrelevant components.

Figure~\ref{fig:exp1_best_auroc} shows that -- except for category \say{grid} -- deeper layers tend to be more informative for \gls{ad} than shallower ones, being capable of achieving perfect score (or very close to it).
This finding contradicts previous results because, when using other strategies (or not using dimension reduction), the deepest layers tend to show a drop in performance (e.g. categories \say{capsule}, \say{pill}, \say{screw}, \say{toothbrush}).
This is believed to happen due to a stronger bias in the deepest layers towards the pre-training task, but our results contradict this explanation.

\paragraph{Bottom-Up vs. Top-Down}
Most scenarios show that the two eigencomponent selection modes, \gls{bottomup} and \say{topdown}, behave similarly.
However, \gls{bottomup} tends to achieve better results with longer plateaus, suggesting that selecting the best eigencomponents leads to better results than removing the spurious ones.

Finally, some exceptional cases were observed and pointed out in Appendix~\ref{sec:exp1_all_graphs}.

\section{Experiment 2: generalization per anomaly type}\label{sec:exp2-peranomtype}

\subsection{Experiment 2 Setup}

Experiment 2 consists of segregating the anomaly types in $\wvecgreedy$ and $\wveceval$. 
Unlike the first experiment, where component selection is based on all anomaly types, this experiment specifically ranks the candidates (see Algorithms~\ref{alg:greedy-bottom-up} and~\ref{alg:greedy-top-down}) using a single anomaly type. 
The evaluation, however, encompasses all the other anomaly types, while both $\wvecgreedy$ and $\wveceval$ use all the normal images from the test set.

This experiment enables us to assess how well the component selection generalizes from a single anomaly type. 
By studying the algorithm's performance under this setup, we can gain deeper insights into the model's ability to extrapolate the learned patterns from one anomaly type to others. 

Notice that category \say{toothbrush} is not in the results of this experiment as it only has one anomaly type.

\subsection{Experiment 2 Results}

\begin{figure}[!h]
    \centering
    \begin{subfigure}[b]{\linewidth}
        \centering
        \adjincludegraphics[
            width=\textwidth,
            trim={{0.25\width} {0.775\height} {0.50\width} {0.155\height}},clip,
            keepaspectratio,
        ]{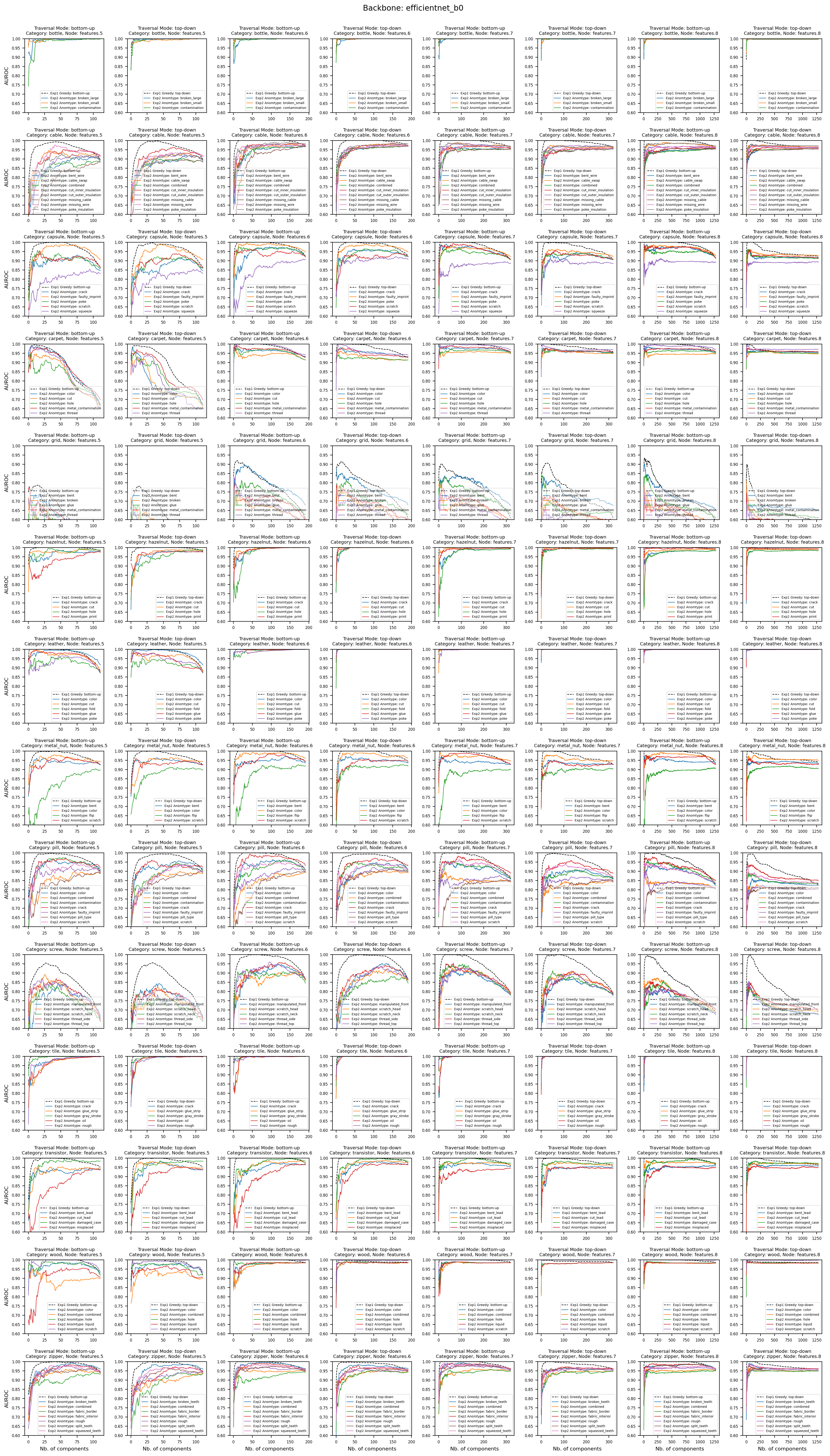}
        \caption{capsule, \features{6}}
        \label{fig:exp2_selection__capsule_f6}
    \end{subfigure}
    \hfill
    \begin{subfigure}[b]{\linewidth}
        \centering
        \adjincludegraphics[
            width=\textwidth,
            trim={{0.50\width} {0.075\height} {0.25\width} {0.855\height}},clip,
            keepaspectratio,
        ]{figures/exp2_per_anomtype.png}
        \caption{wood, \features{7}}
        \label{fig:exp2_selection__wood_f7}
    \end{subfigure}
    \caption{
        Experiment 2: selection of representative cases.
    }
    \label{fig:exp2_selection}
\end{figure}

The results of Experiment 2 for all scenarios can be found in Appendix~\ref{sec:exp2_all_graphs}, and Figure~\ref{fig:exp2_selection} shows a selection of representative cases.
The results from nodes from \features{0} up to \features{4} have been omitted because their \textit{achievable} performances (see Experiment 1, Section~\ref{sec:exp1-overfit}) are generally worse and they have less stable behavior than nodes from \features{5} up to \features{8}. 

All the possible configurations for a given category (named after the anomaly type used in $\wvecgreedy$) are compared with the respective curve from Experiment 1.
Notice that each curve has a different $\wveceval$, so they do \textit{not} have the same performance at the point $\numdimred = \numdim$.

\subsection{Experiment 2 Discussion}



In Figure~\ref{fig:exp2_selection__capsule_f6}, we observe that Experiment 1 outperforms all the curves based on a single anomaly type in both modes (\gls{bottomup} and \gls{topdown}). 
In other words, the greedy eigencomponent selection struggles to generalize well to unseen anomalies.
Besides, its adaptability across diverse categories lacks consistency.

While Experiment 1 shows its possible to obtain a perfect classifier with less than 30 eigencomponents -- and adding up to other 70 eigencomponents does not hurt the performance -- none of the single-anomaly-type runs were capable of ever reaching such performance.
As shown in Appendix~\ref{sec:exp2_all_graphs}, most scenarios have this behavior with more or less cross-anomaly type variability (e.g. category \say{capsule} has a stronger dependency on the anomaly type than the category \say{cable}).


Figure~\ref{fig:exp2_selection__wood_f7} shows a more stable behavior in the sense that all the anomaly types have nearly the same curve.
The greedy runs in Experiment 2 are comparable and more often better than \gls{pca} and \gls{npca}, which comes without surprise due to the supervision used in the former. 
However, even in such cases, results from Experiment 2 generally fail to achieve the same level of performance seen in Experiment 1.


\paragraph{Bottom-Up vs. Top-Down}

\gls{bottomup} often reaches better maximum performance with lower k, while \gls{topdown} is more stable at keeping the baseline performance (no dimension reduction) and shows a more consistent behavior.
Categories \say{carpet} and \say{zipper} (node \features{7} in particular) are good examples of such contrast.
Other examples include categories  \say{hazelnut}, \say{leather}, \say{transistor}, and \say{wood}.

\section{Experiment 3: generalization with fixed number of images}\label{sec:exp3-fixednumimgs}

\subsection{Experiment 3 Setup}

Instead of basing the choice of components on the whole test set performance (Experiment 1) or on a single anomaly type (Experiment 2), we now establish a minimum number of anomalous images $\minnumimg$ in $\wvecgreedy$ and randomly select images from all anomaly types.
Each scenario is repeated 5 times with a different seed for the greedy-evaluation split ($\wvecgreedy$/$\wveceval$).

To further clarify, the $\wvecgreedy$ set includes anomalous images with at least our predefined minimum $\minnumimg$, chosen proportionally from each anomaly type within a category. 
The remaining anomalous images constitute $\wveceval$, while both $\wvecgreedy$ and $\wveceval$ use all the normal images from the test set.
More details in Table~\ref{tab:num_images} in Appendix~\ref{sec:mvtec-num-images}.

This approach lends us the flexibility to learn and evaluate on diverse data sets, without bias towards any particular anomaly type.


\subsection{Experiment 3 Results}

\begin{figure}[!t]
    \centering
    \begin{subfigure}[b]{\linewidth}
        \centering
        \adjincludegraphics[
            width=\textwidth,
            trim={{0.75\width} {0.790\height} {0.011\width} {0.143\height}},clip,
            keepaspectratio,
        ]{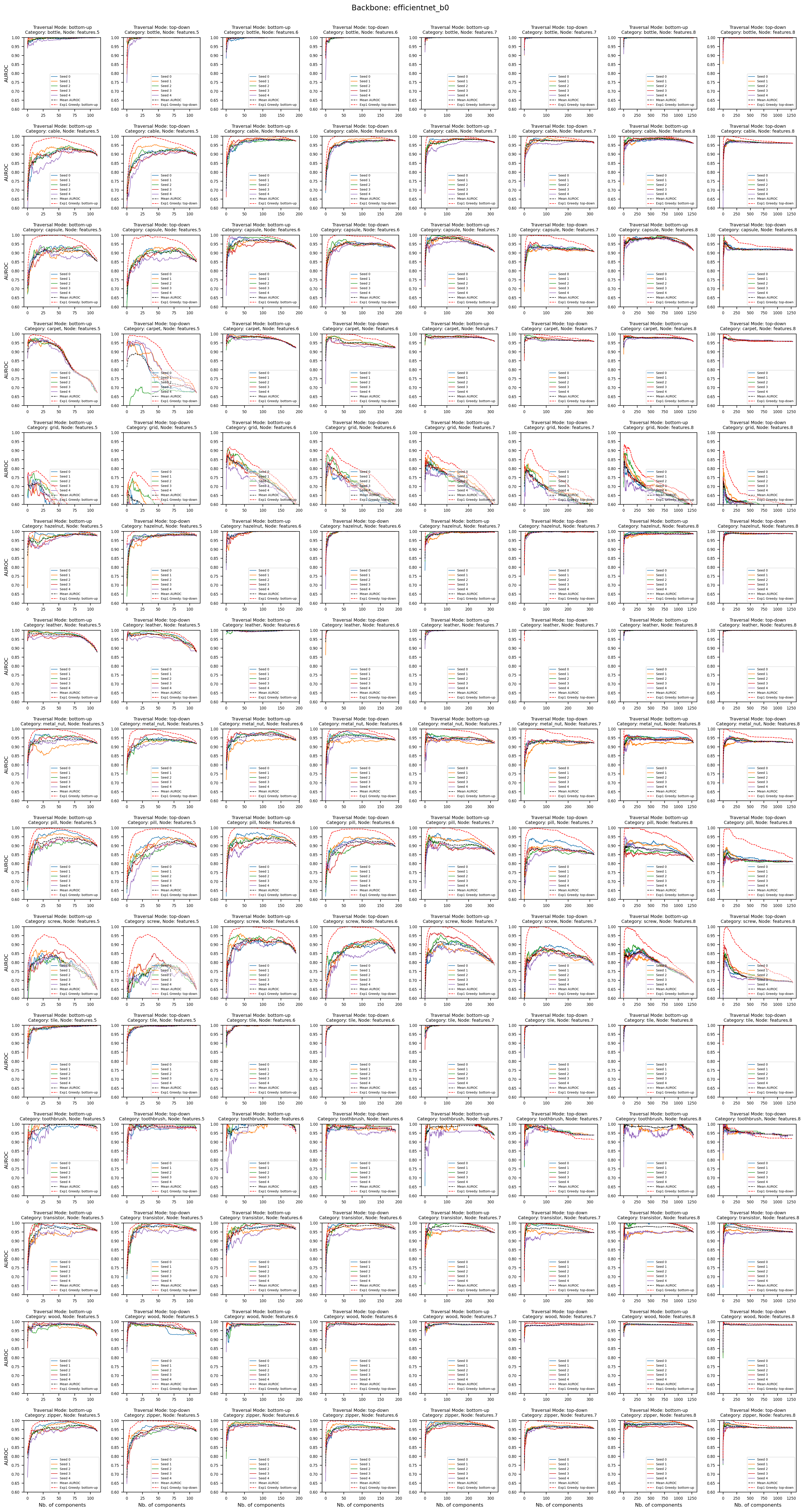}
        \caption{capsule, \features{8}}
        \label{fig:exp3_selection__capsule_f8}
    \end{subfigure}
    \hfill
    \begin{subfigure}[b]{\linewidth}
        \centering
        \adjincludegraphics[
            width=\textwidth,
            trim={{0.50\width} {0.465\height} {0.25\width} {0.470\height}},clip,  
            keepaspectratio,
        ]{figures/exp3_mean_line.png}
        \caption{metal nut, \features{7}}
        \label{fig:exp3_selection__metanut_f7}
    \end{subfigure}
    \caption{
        Experiment 3: selection of representative cases.
    }
    \label{fig:exp3_selection}
\end{figure}

The results of Experiment 3 for all scenarios can be found in Appendix~\ref{sec:exp3_all_graphs}, and Figure~\ref{fig:exp3_selection} shows a selection of representative cases with the $\numdimred$-vs-\gls{auroc} curves from all seeds separately, their cross-seed mean curve, and the curve from Experiment 1 for reference. 
The results from nodes from \features{0} up to \features{4} have been omitted because their \textit{achievable} performances (see Experiment 1, Section~\ref{sec:exp1-overfit}) are generally worse and they have less stable behavior than nodes from \features{5} up to \features{8}. 

Figure~\ref{fig:exp3_selection} presents two representative cases, aiming to illustrate distinct scenarios when employing a fixed number of images. 


\subsection{Experiment 3 Discussion}

Compared to Experiment 2, Experiment 3 shows, as expected, slightly better results with less variance across runs of a same scenario, which is expected because $\wvecgreedy$ is not biased towards a single anomaly type -- two counter examples are worth noting: categories \say{pill} and \say{transistor}.
Still, a similar pattern often arises: while the curve from Experiment 1 reaches 100\% \gls{auroc}, the others fail to avoid bad components.

Figure~\ref{fig:exp3_selection__metanut_f7} shows a noticeable pattern in Experiment 3. 
While Experiment 1 reveals a rather important margin for improvement (relative to the baseline without dimension reduction), the ability to generalize with reduced amount of data is very limited, and the discrepancy is usually bigger for with the \gls{bottomup} mode.
Again, the greedy selection fails to avoid bad components, although performances are generally better than \gls{pca} and \gls{npca}, which, again, is not surprising because the latter have no supervision at all.
Other noticeable examples include categories \say{capsule}, \say{carpet}, \say{metal nut}, \say{pill}, and \say{screw}.

Figure~\ref{fig:exp3_selection__capsule_f8} shows an encouraging example where the \gls{bottomup} approach is more successful.
Most runs achieved substantial performance improvement relative to the baseline (no dimension reduction) with low variability.
Category \say{carpet} with node \features{6} and category \say{leather} with node \features{5} also represent well such behavior.

In cases where the baseline typically has more than 95\%, this is often generally the case as well.
Although the relative improvement is not as prominent (baseline is already high), the dimension reduction -- without loss or gain of performance -- is considerable.
Some examples include categories \say{cable} and \say{zipper} with nodes from \features{6} to \features{8}, and categories \say{carpet} and \say{hazelnut} with node \features{8}.



\paragraph{Bottom-Up vs. Top-Down}

The same pattern observed in Experiment 2 is seen here: \gls{bottomup} is more embedding-size-efficient, while \gls{topdown} manages to only keep the baseline performance (no dimension reduction).
In Experiment 3, however, the two modes don't differ as much, making the \gls{topdown} a safer choice.
Examples of this can be seen in categories \say{capsule} and \say{hazelnut} (nodes from \features{6} to \features{8}).

\section{Other analyses}

Despite unrealistic, Experiment 1 shows surprisingly efficient results, so we additionally present more detailed analyses of its results in Appendix~\ref{sec:index_index} and Appendix~\ref{sec:redundancy_noise}.

Although the variance in a vector space is commonly used as a reference signal in many data techniques, like \gls{pca}-based methods, Appendix~\ref{sec:index_index} reveals that there is no connection between eigencomponent-wise variance and \gls{ad} suitability.
Otherwise said, the eigencomponents with the largest or smallest variance do not necessarily discriminate normal from anomalous data better than one, contradicting the core premises of \gls{pca} and \gls{npca} respectively.
Appendix~\ref{sec:index_index} shows that, in fact, the variance of the eigencomponents present in the rise regime (see Appendix~\ref{fig:regimes} for more details) does not have any pattern in the high-performance scenarios from Experiment 1. 

Finally, Appendix~\ref{sec:redundancy_noise} shows two simulated experiments investigating the nature of the three regimes observed, specially with the \gls{bottomup} mode, in most high-performance scenarios: rise, plateau, drop.
The results suggest that the eigencomponents in the plateau regime behave like redundant synthetic data, actually with more stable behavior than the latter.
The eigencomponents in the drop regime, however, have spurious features that do not discriminate the normal from the anomalous class -- in fact provoking a faster performance drop than pure noise. 


\section{Conclusion}

The paper presents three experiments evaluating a novel dimension reduction for anomaly detection (AD) combining eigendecomposition of the covariance matrix and a greedy tree search algorithm. 
The first experiment intentionally overfits the test set to measure the algorithm's potential and compare it with previous approaches showing that it is possible to achieves high performance with small embeddings, outperforming previous approaches, namely \gls{pca} and \gls{npca}.
Our proposed analysis demonstrated surprisingly results contradicting previous results, and additional analyses revealed that the variance in a vector space does not directly correlate with anomaly detection performance. 

However, the second and third experiments, exploring its generalization capacity, reveal that the algorithm struggles to generalize well to unseen anomalies.
Although the features extracted from the studied \gls{cnn} contain redundant and spurious information (to discriminate normal from anomalous instances), identifying them with reduced amount of data remains challenging.

We hypothesize that the struggles observed in Experiment 2 and Experiment 3 could potentially be mitigated by exploring different criteria or metrics for evaluating the quality of eigencomponents during the dimension reduction process. 
Further research and investigations into novel heuristics could better leverage the valuable insights observed in this work, thus advancing the effectiveness of the proposed approach for anomaly detection.

\clearpage
\nocite{*}
{\small
\bibliographystyle{ieee_fullname}
\bibliography{main}
}


\newpage
\appendix
\onecolumn
\section{Experiment 1: test set overfit}\label{sec:exp1_all_graphs}

Figure~\ref{fig:exp1_all_graphs} represents the \gls{auroc} performance for component-wise analysis ($\numdimred$ vs. \gls{auroc} curves), as described in the Section~\ref{sec:exp1-overfit}.
It provides a visual depiction of how individual components within nodes perform across different categories. 
Each row in the graph represents a category, while each column represents a node.

The x-axis of the graph corresponds to the number of components $\numdimred$ (i.e. "how much is $\numdim$ reduced to?"). 
In this experiment, we rank all the eigencomponents of the $\empcovmat$.
The Y-axis indicates the \gls{auroc} value, which measures the predictive performance of the components.

The graph includes separate curves that represent different approaches used in the analysis. 
It shows the \gls{auroc} performance of the greedy search for both traversal modes (Bottom-up and Top-down) compared to the results obtained from other dimension reduction strategies, including \gls{pca},\gls{npca}, and the supspace "$\left[ \Phi_{2}, \Phi_{3} \right]$" introduced in \cite{lin_deep_2022}.

Figure~\ref{fig:exp1_zoom_all_graphs} is part of the same experiment but the X-axis is restricted to the $\numdimred \leq 40$ components (out of $\numdim$), the Y-axis remains the same with the respective \gls{auroc}s. 
These graphs also include the \gls{pca}, \gls{npca} results with a horizontal dashed line marking the highest \gls{auroc} for the respective method.
We also mark the \gls{auroc} performance with no dimension reduction ($\numdim$ components) with a  horizontal dashed line to highlight the importance of the component selection. 

\paragraph{Exceptional case 1: toothbrush is easier}

Our analysis revealed an exceptional case in the category of \say{toothbrush}, where all the nodes achieved nearly-maximum performance.
This outcome suggests that detecting anomalies in this category is comparatively easier than in other categories.
Although, it must be noted that this category has the smallest test set with only 30 anomalous images while others have nearly twice or three times as many, and it only has one anomaly type while others such as \say{pill}, \say{screw}, and \say{zipper} have between 5 and 7 anomaly types.
See Table~\ref{tab:num_images} in Appendix~\ref{sec:mvtec-num-images}.

\paragraph{Exceptional case 2: very low-dimensional categories}

The greedy eigencomponent selection achieves particularly impressive results in categories \say{wood}, \say{leather}, \say{tile}, and \say{bottle}.
A perfect score (100\% \gls{auroc}) is consistently achieved with less than 5 eigencomponents in several nodes.

\begin{figure}[!b]
    \centering
    \adjincludegraphics[
        width=\textwidth,
        trim={0 {0.66\height} 0 0},clip,
        keepaspectratio,
    ]{figures/full_depth.png}
    \phantomcaption
\end{figure}
\begin{figure}[!t]
    \ContinuedFloat
    \centering
    \adjincludegraphics[
        width=\textwidth,
        trim={0 0 0 {0.34\height}},clip,
        keepaspectratio,
    ]{figures/full_depth.png}
    \caption{Experiment 1: all plots.}
    \label{fig:exp1_all_graphs}
\end{figure}

\clearpage
\begin{figure*}[!h]
    \centering
    \adjincludegraphics[
        width=\textwidth,keepaspectratio,
        trim={0 {0.66\height} 0 0},clip,
    ]{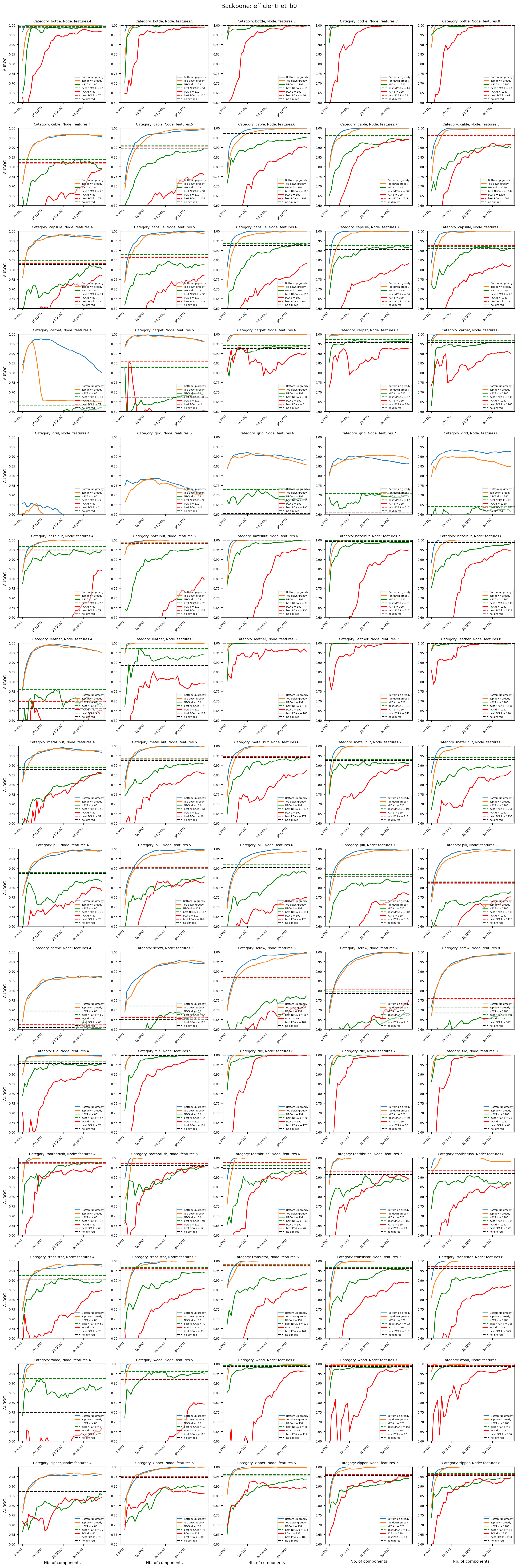}
    \phantomcaption
\end{figure*}
\begin{figure*}[!h]
    \ContinuedFloat
    \centering
    \adjincludegraphics[
        width=\textwidth,keepaspectratio,
        trim={0 {0.334\height} 0 {0.34\height}},clip,
    ]{figures/zoom.png}
    \phantomcaption
\end{figure*}
\begin{figure*}[!h]
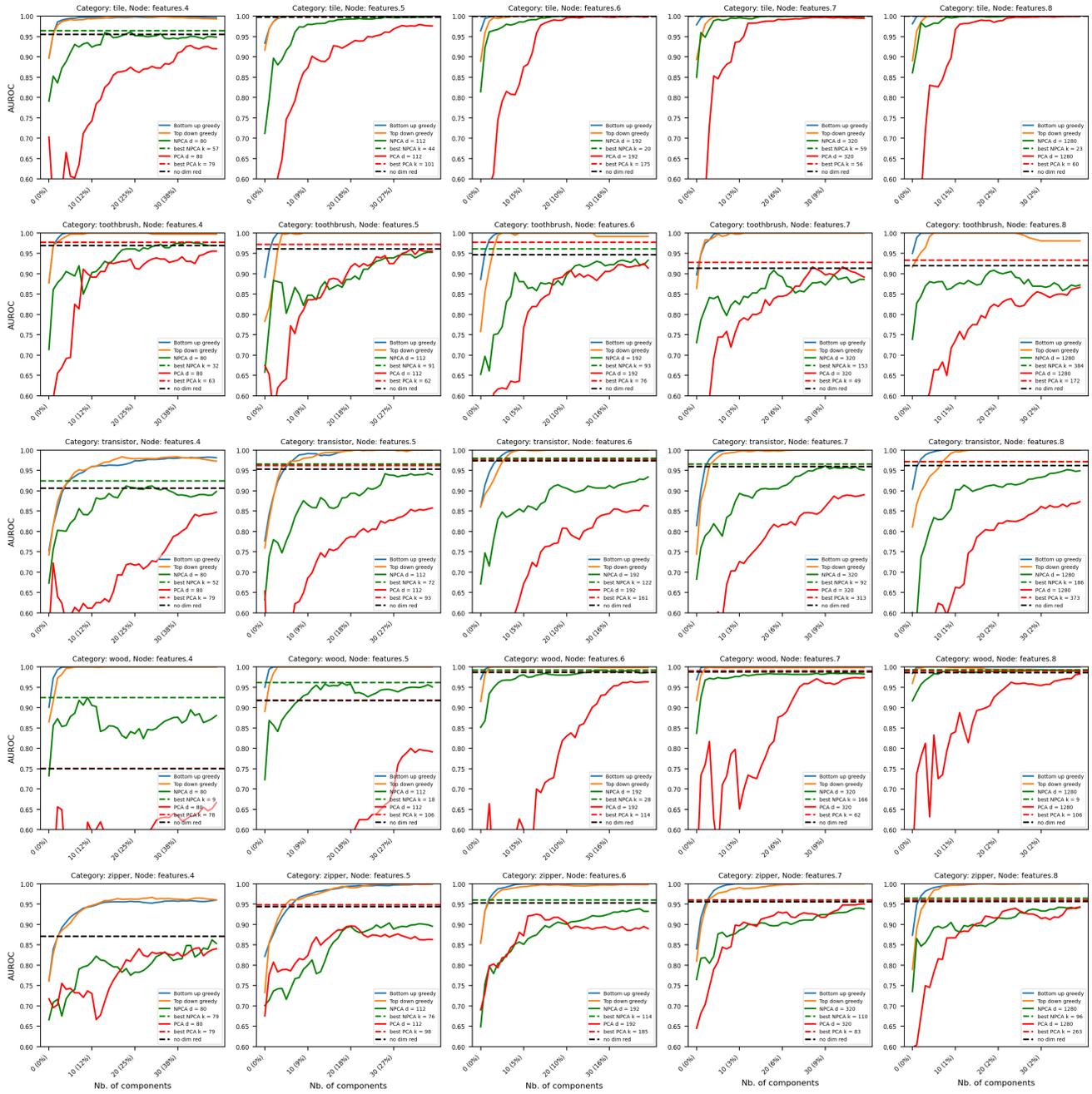

    \ContinuedFloat
    \centering
    \adjincludegraphics[
        width=\textwidth,keepaspectratio,
        trim={0 0 0 {0.668\height}},clip,
    ]{figures/zoom.png}
    \caption{
        Experiment 1: all plots. 
        Zoom on the X-axis $\numdimred \leq 40$ components.
    }
    \label{fig:exp1_zoom_all_graphs}
\end{figure*}

\clearpage
\section{Experiment 2: generalization per anomaly type}\label{sec:exp2_all_graphs}

Figure~\ref{fig:exp2_all_graphs} shows all the scenarios, as discussed in Section~\ref{sec:exp2-peranomtype}, of the results obtained from Experiment 2. 
These plots grant critical insights into the performance of the various anomaly types, along with the influence of the component selection strategy on the generalization process. 
Figure~\ref{fig:exp2_all_graphs} also includes a juxtaposition of the results from Experiment 2 along with the results from Experiment 1 for reference. 

The line plots illustrate the number of eigencomponents on the X-axis and the corresponding \gls{auroc} values on the Y-axis.
Notice that each curve corresponds to a different $\wvecgreedy$ / $\wveceval$ splits, so they do not match the same performance at the point $\numdimred = \numdim$.

\begin{figure}[!b]
    \centering
    \adjincludegraphics[
        width=\textwidth,
        trim={0 {0.425\height} 0 0},clip,
        keepaspectratio,
    ]{figures/exp2_per_anomtype.png}
    \phantomcaption
\end{figure}
\begin{figure}[!h]
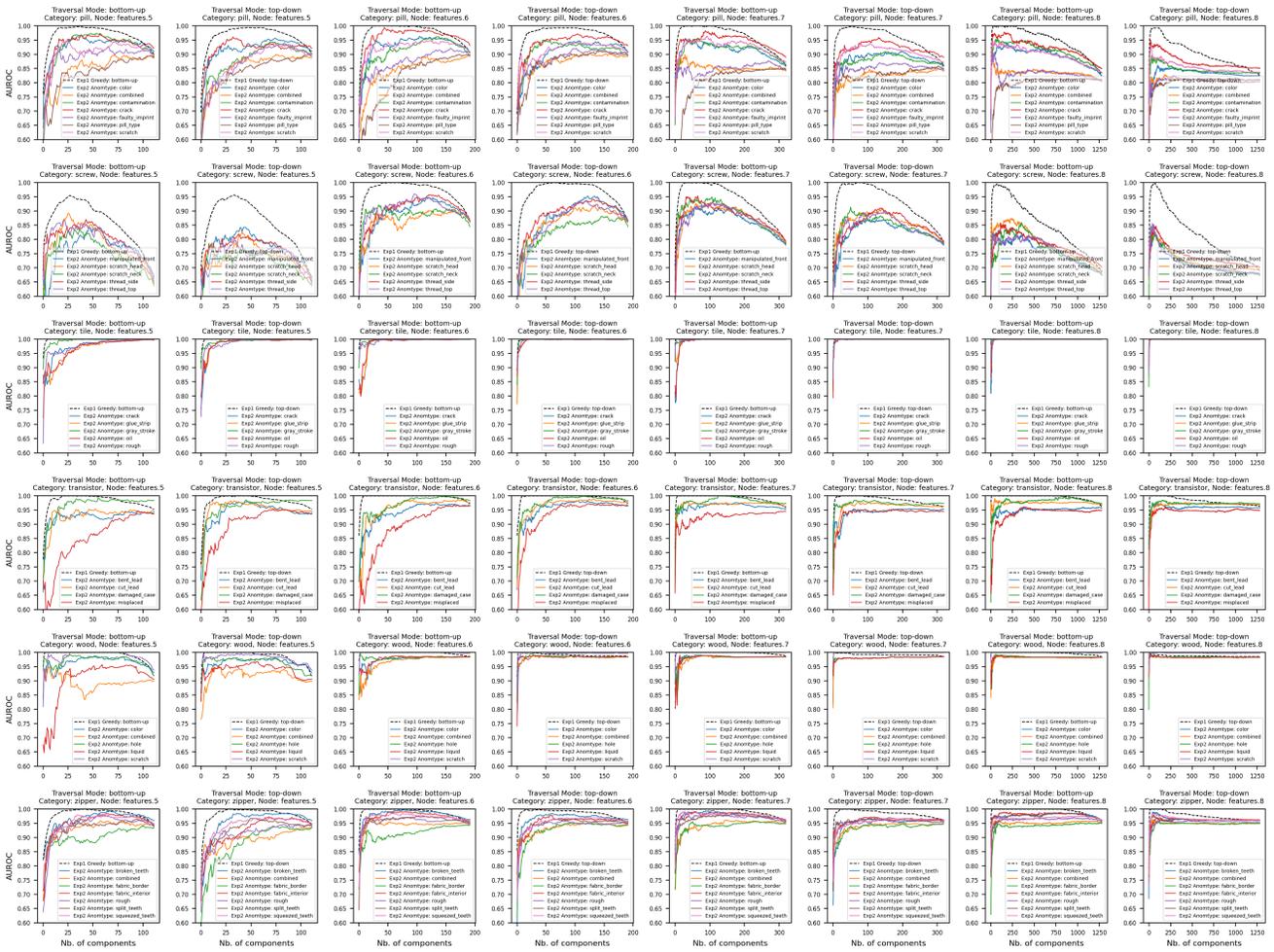

    \ContinuedFloat
    \centering
    \adjincludegraphics[
        width=\textwidth,
        trim={0 0 0 {0.575\height}},clip,
        keepaspectratio,
    ]{figures/exp2_per_anomtype.png}
    \caption{Experiment 2: all plots.}
    \label{fig:exp2_all_graphs}
\end{figure}

\clearpage
\section{Experiment 3: generalization with fixed number of images}\label{sec:exp3_all_graphs}

Figure~\ref{fig:exp3_all_graphs} shows all the scenarios, as discussed in Section~\ref{sec:exp3-fixednumimgs}, of the results obtained from Experiment 3. 
The figure also includes a comparison of the results from Experiment 3 with both Bottom-Up and Top-Down traversal modes employed in Experiment 1.
The line plots in the figure depict the number of eigencomponents on the X-axis and the corresponding AUROC values on the Y-axis. 

These plots show the performance of the various seeds along with their (cross-seed) mean performance and the curve from Experiment 1 for reference, providing valuable insights into the generalization capacity of the greedy eigencomponent selection.

\begin{figure}[!b]
    \centering
    \adjincludegraphics[
        width=\textwidth,
        trim={0 {0.463\height} 0 0},clip,
        keepaspectratio,
    ]{figures/exp3_mean_line.png}
    \phantomcaption
\end{figure}
\begin{figure}[!h]
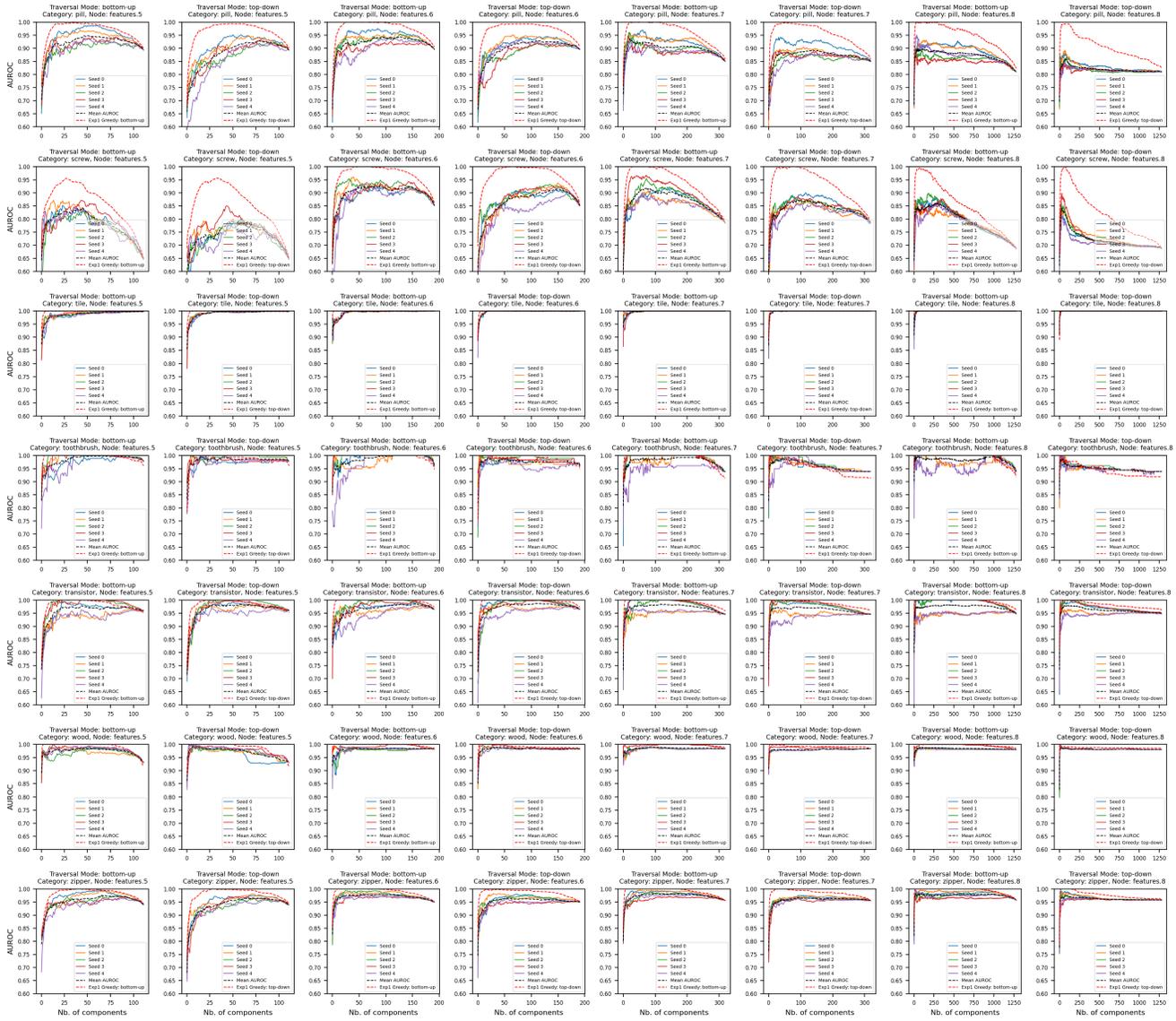

    \ContinuedFloat
    \centering
    \adjincludegraphics[
        width=\textwidth,
        trim={0 0 0 {0.540\height}},clip,
        keepaspectratio,
    ]{figures/exp3_mean_line.png}
    \caption{Experiment 3: all plots.}
    \label{fig:exp3_all_graphs}
\end{figure}

\clearpage
\section{Regimes}\label{sec:regimes}

\begin{figure}[!b]
    \centering
    \includegraphics[
        width=\textwidth,
        keepaspectratio,
    ]{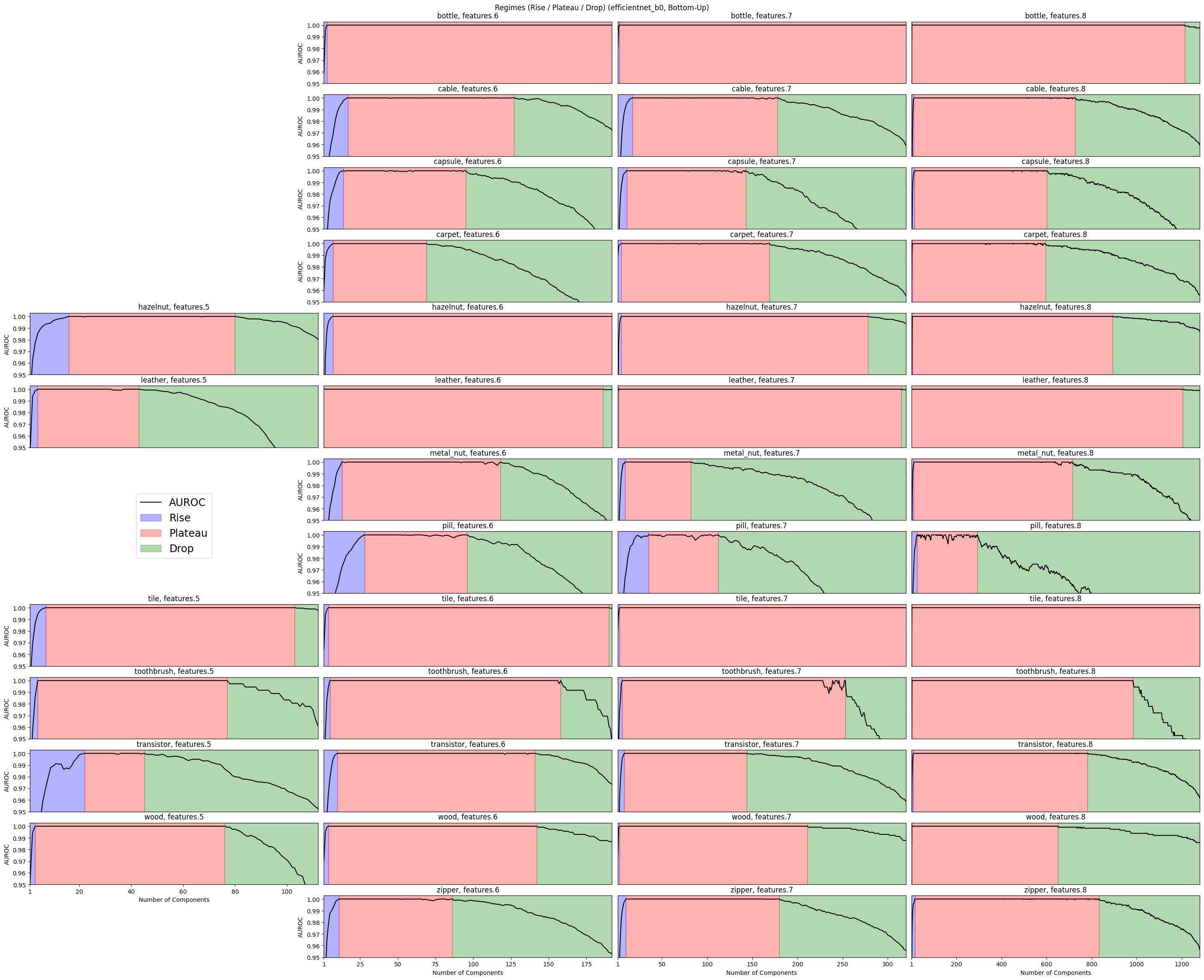}
    \caption{Experiment 1: eigencomponents regimes.}
    \label{fig:regimes}
\end{figure}

Figure~\ref{fig:regimes} is derived from the results of Experiment 1, where we distinguish three main regimes from the greedy eigencomponent selection of the \gls{bottomup} traversal mode.
We designate the \say{Rise} in blue, the \say{Plateau} in red, and the \say{Drop} in green. The use of distinct colors provides a better visualisation and differentiation of these three regimes.

\paragraph{We selected representative cases and based this choice on a certain criteria:} we included cases were \gls{auroc} reaches the score of 1, and deliberately focused on showing only $\geq \features{5}$ because starting from this node we can clearly observe different regimes and because they achive the best performance. 
We also excluded categories such as \say{screw} and \say{grid} as their results didn't align with our regime analysis, meaning the results made it hard to distinguish these three regimes. 

\paragraph{The Rise regime (blue):} This phase represents the initial selection of eigencomponents where each new addition significantly boosts the performance. 
 
\paragraph{The Plateau regime (red):} In this phase the addition of more eigencomponents doesn't significantly improve or degrade the performance. In fact, for most cases with only several exceptions we observe almost all the time a 100\% \gls{auroc}. 

\paragraph{The Drop regime (green):} Finally, the 'Drop' regime represents the phase where adding more eigencomponents starts to degrade the performance. This can happen due to the incorporation of noisy, irrelevant, or redundant components which don't contribute positively and are simply bad components. 

\paragraph{}
The most interesting cases would be \say{cable}, \say{capsule}, \say{pill}, \say{transistor}, and \say{zipper} with the nodes \features{6} and \features{7} because we can see clearly all the three regimes and utilize its eigencomponents for further analysis mentioned in Sections~\ref{sec:k_at_max_auroc} and ~\ref{sec:index_index}. 


\clearpage
\section{Minimal Number of Dimensions at Maximum AUROC}\label{sec:k_at_max_auroc}

\begin{figure}[!h]
    \centering
    \includegraphics[
        width=0.90\textwidth,
        keepaspectratio,
    ]{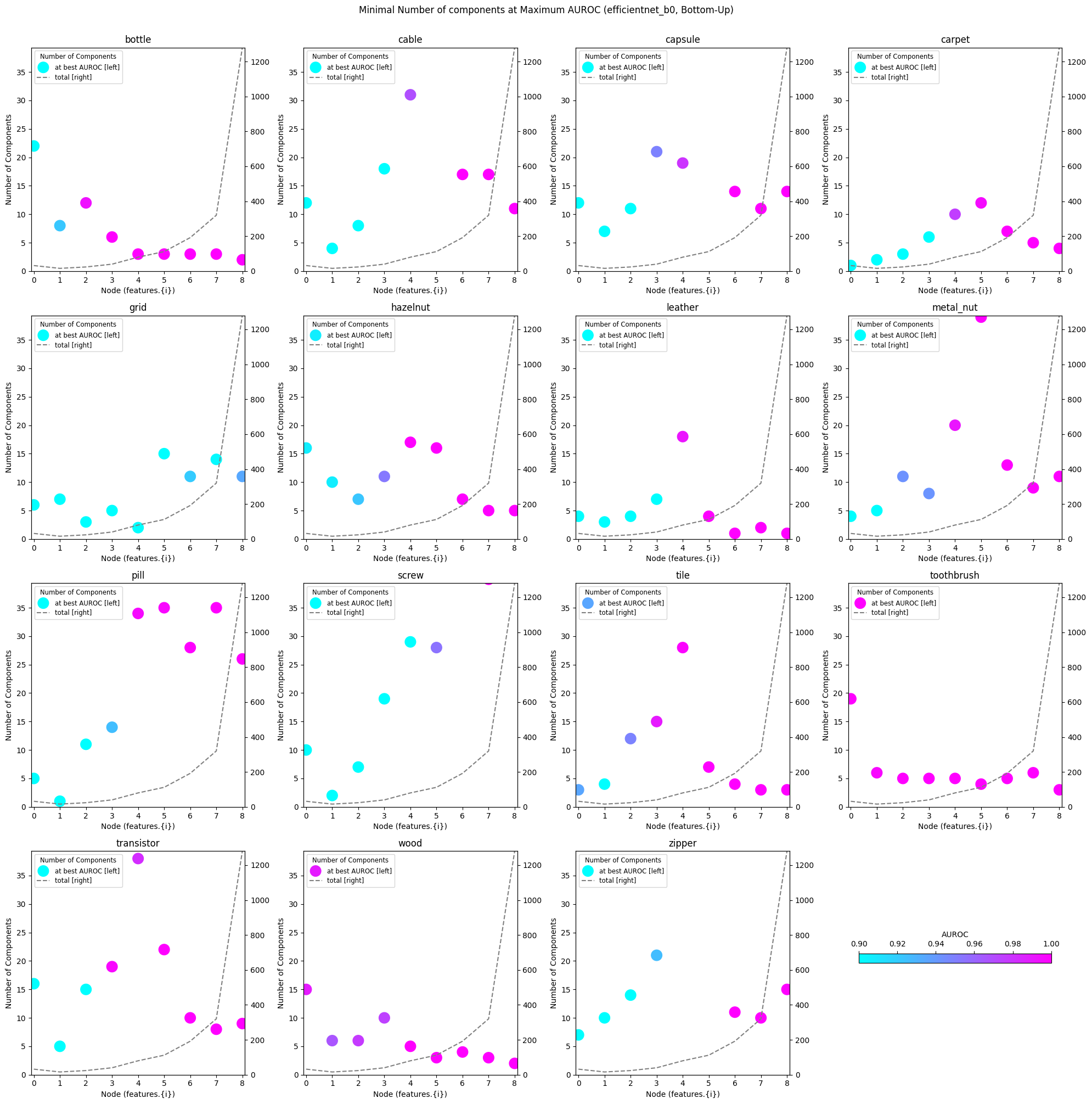}
    \caption{Experiment 1: minimal number of (reduced) dimensions $\numdimred$ at maximum AUROC.}
    \label{fig:k_at_max_auroc}
\end{figure}

Figure~\ref{fig:k_at_max_auroc} shows an analysis of the optimal dimension reduction size for all the scenarios in Experiment 1 (Section~\ref{sec:exp1-overfit}).
We select minimal number of dimensions $\numdimred$ such that its corresponding \gls{auroc} is maximal -- it corresponds to the left most point of the plateau regime (see Figure~\ref{fig:regimes} in the Appendix~\ref{sec:regimes}).

The X-axis corresponds to the node depth in \gls{effnetb0} and the Y-axis shows its corresponding optimal number of components $\numdimred$ (dots, scaled on the left) and the dashed line (scaled on the right) shows the original feature vector size $\numdim$.  
The marker color represent a point's corresponding \gls{auroc} scaled from $0.90$ to $1.00$ (light blue to pink).

The optimal number of components (left y-axis) of high-performance models (pink markers) has a negative trend relative to the node depth (deeper nodes implicate less components), while the original embedding size (right y-axis) is bigger.
In other words, higher dimensional embeddings tend to be capable of encoding the normality of the images in lower dimensional subspaces.

\clearpage
\section{Are the best components from the smallest or largest eigenvalues? Both.}\label{sec:index_index}


We analyse if the order that components show up in the bottom-up component selection relates to \gls{pca} or \gls{npca}.
Figure~\ref{fig:index_index} shows the step index vs. the component index of all the runs with \gls{effnetb0} using the bottom-up strategy.

The x-axis is the step index, which corresponds to the depth of the search tree in greedy, so it represents, from left to right, the order that the eigencomponents were added. 
The y-axis is the component index, which corresponds to the eigenvalues order, so it represents, from bottom to top, the smallest to highest eigenvalues.
\gls{pca}'s graph would be a line with slope $-1$ (first component is the largest, last component is the smallest), and \gls{npca}'s would be a line with slope $1$ (first component is the smallest, last component is the largest).

The most relevant nodes are from \features{5} to \features{8} because they achieve the best performances (see Figures~\ref{fig:exp1_all_graphs} and~\ref{fig:exp1_best_auroc}).
The most relevant components are generally the first 10 to 30 ones (left most part of each plot; see Figure~\ref{fig:k_at_max_auroc}) because that generally corresponds to the \say{rise} regime (see Section~\ref{sec:regimes}).
The cases with most visible regimes are plotted with the same regime colors used in Figure~\ref{fig:regimes}.

Despite some exceptionally structured cases like (some nodes from) categories bottle and tile, which are similar to the \gls{pca} component sorting, most plots are disordered and seemingly random.
This shows there is no relation between eigenvalue magnitude (i.e. the amount of variance encoded in an eigencomponent) and utility for anomaly detection, which was implicitely assumed in previous works using \gls{pca}, \gls{npca}, and in \cite{lin_deep_2022}.
Notice this lack of entanglement between the two is particularly visible at the first (therefore most useful) components. 

\begin{figure}[!b]
    \centering
    \adjincludegraphics[
        width=\textwidth,
        trim={0 {0.595\height} 0 0},clip,
        keepaspectratio,
    ]{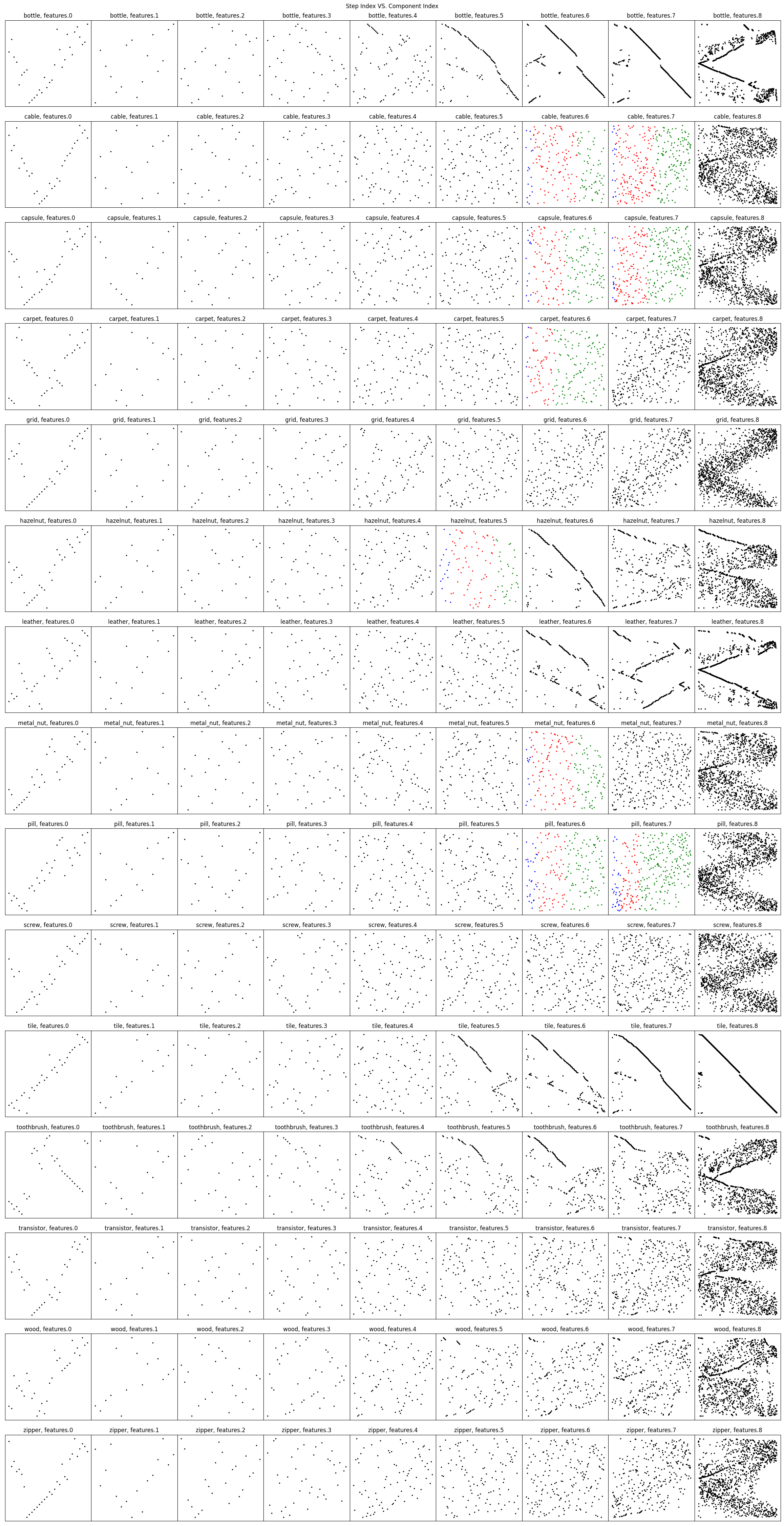}
    \phantomcaption
\end{figure}
\begin{figure}[!h]
    \ContinuedFloat
    \centering
    \adjincludegraphics[
        width=\textwidth,
        trim={0 0 0 {0.405\height}},clip,
        keepaspectratio,
    ]{figures/index_index.png}
    \caption{Step index vs. component index. }
    \label{fig:index_index}
\end{figure}

\clearpage
\section{Are the eigencomponents from the plateau and drop regions redundant? Or noisy?}\label{sec:redundancy_noise}


\newcommand{\startingk}{\numdimred^{\prime}}

\begin{figure}[!hb]
    \centering
    \includegraphics[
        width=.95\textwidth,
        keepaspectratio,
    ]{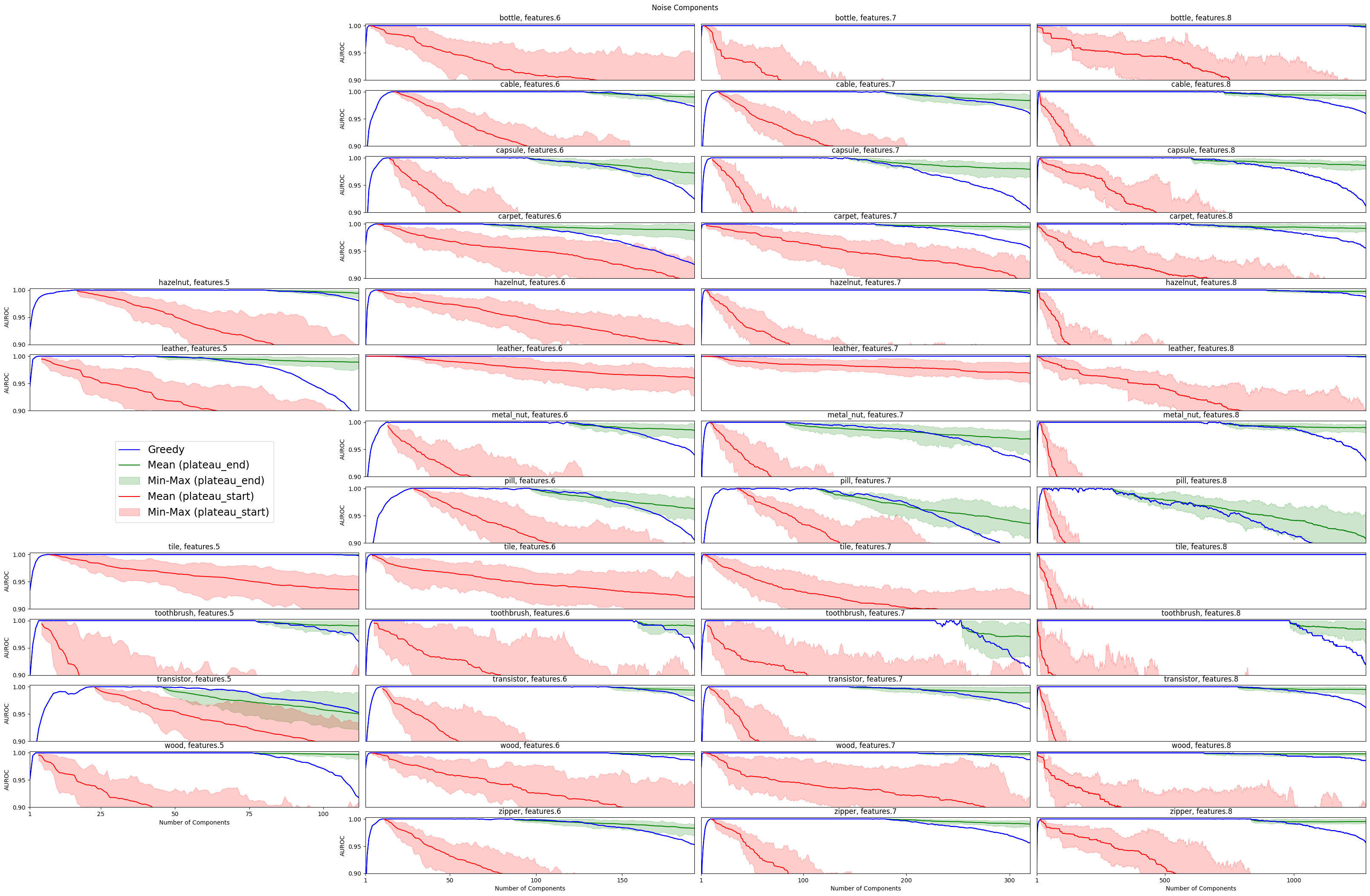}
    \caption{
        Simulated performance with \textbf{noise} signal. 
        Red/green: replacement starts at the start/end of the "plateau" regime.  
    }
    \label{fig:noise}
\end{figure}

\begin{figure}[!hb]
    \centering
    \includegraphics[
        width=.95\textwidth,
        keepaspectratio,
    ]{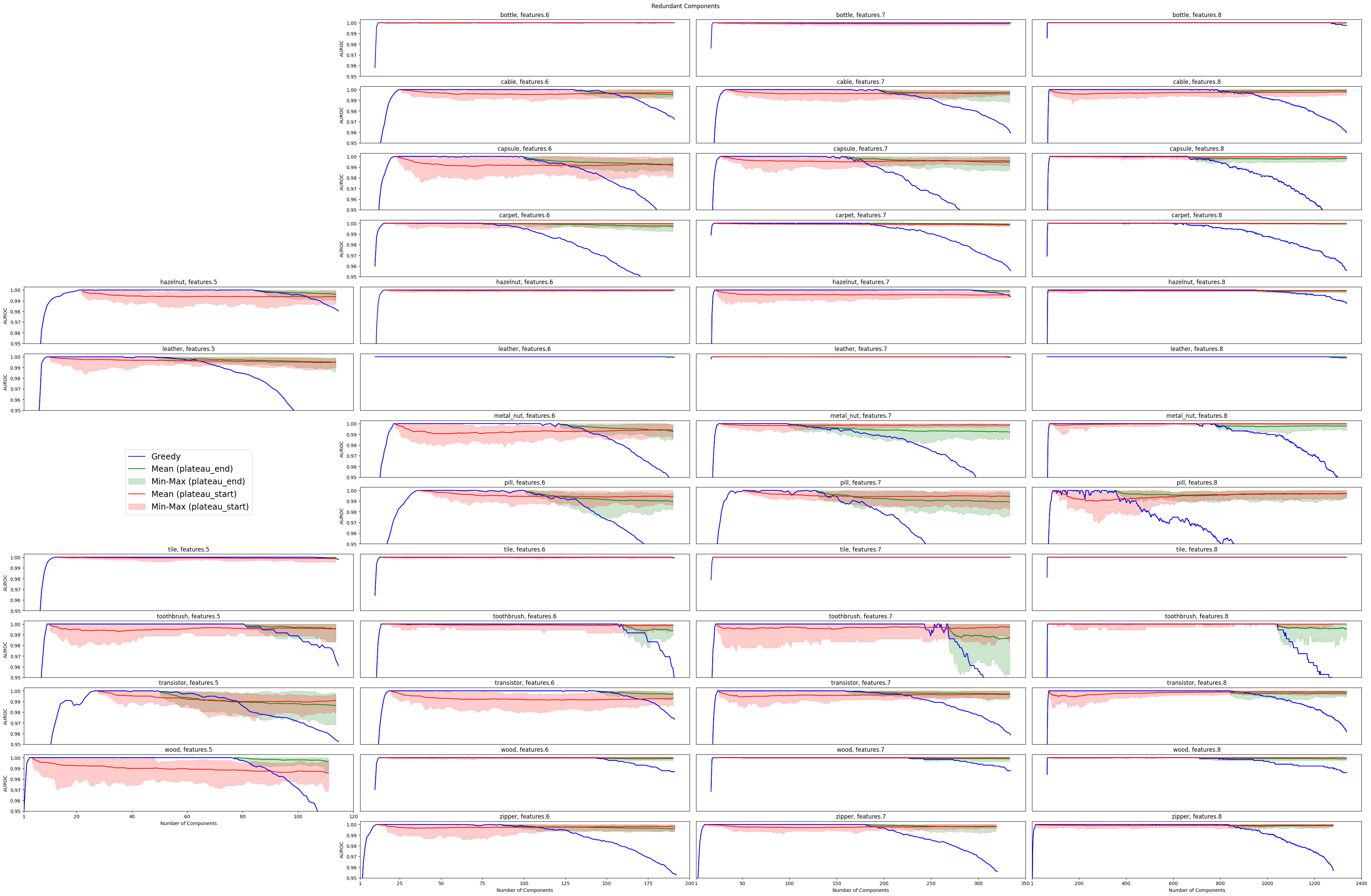}
    \caption{
        Simulated performance with \textbf{redundant} signal. 
        Red/green: replacement starts at the start/end of the "plateau" regime.  
    }
    \label{fig:redundancy}
\end{figure}

Can one say that the components selected after the \say{rise} regime (cf. definition of regimes in Section~\ref{sec:regimes}) are contain redundant, noisy, or even spurious information?
Figures~\ref{fig:redundancy} and~\ref{fig:noise} present the outcomes of two simulations aimed at addressing these questions.

The $\numdim$ eigencomponents of a Gaussian model are sorted as they were yielded by the results of the bottom-up greedy search using the full test set (i.e. overfit), so a dimension reduction with $\numdimred$ components corresponds to the $\numdimred$-th step (or depth) of the search tree traversal.
This order of eigencomponents is interpreted as \say{from the most useful to the least useful (or most harmful)} relative to the \gls{ad} performance.

Starting at a position $\startingk \in \{ 1, \dots, \numdim \}$ of this list, the eigencomponents before it are retained, the rest is discarded, then synthetic axes are progressively added to replace the discarded ones.
The respective performance of each simulated dimension reduction (original plus synthetic axes) is recorded and compared to the the original dimension reduction with the same size $k$.
Example: for a feature vector with size $\numdim = 10$, a starting position $\numdimred^{\prime} = 4$, and simulated $\numdimred = 5$, the first $3$ components are from the actual eigendecomposition (according to the greedy search order) and the two last components are synthetic. 

Two starting positions $\startingk$ are considered: the first and the last position of the \say{plateau} regime, which corresponds to retaining, respectively, the \say{rise} components and the \say{rise + plateau} components.
Two types of synthetic signal are considered: noise (Figure~\ref{fig:noise}) and redundant signals (Figure~\ref{fig:redundancy}).
The noise is drawn from a standard normal distribution (all axes are independent).
The redundant signal is a Gaussian random projection\footnote{We use the implementation from scikit-learn (\texttt{sklearn.random\_projection.GaussianRandomProjection}).} of the $\startingk - 1$ retained axes (original eigencomponents) -- in other words, multiple random linear combinations of the existing signal.
Four scenarios, considering the combinations of the aforementioned parameters, are considered tested with the feature vector size $\numdimred \in \{ \startingk, \dots, \numdim \}$.
  
To make a fair comparison between the simulated performance and the original performance, the scale of a synthetic axis at position $i$ is chosen such that its empirical standard deviation on the test set $\hat{\sigma}_{\text{test}}^{(i)}$ equals that of the original component at the position $i$ in the sorted list.
Each scenario is repeated with 30 different random seeds; Figure~\ref{fig:noise} and Figure~\ref{fig:redundancy} show the minimum, average, and maximum performance seen at each dimension reduction size $\numdimred$.

Most cases (category and node combination) show a consistent behavior: compared to the components in the \say{plateau} regime, noise deteriorates the performance, and faster (less synthetic axes) than the redundant signal, which often remains close to the plateau's performance.

These results suggest that the eigencomponents in the plateau regime behave like redundant synthetic data, actually with more stable behavior than the latter.
The eigencomponents in the drop regime, however, have spurious features that do not discriminate the normal from the anomalous class -- in fact provoking a faster performance drop than pure noise.


\clearpage
\section{MVTec-AD Dataset Overview}\label{sec:mvtec-num-images}

The table provides a comprehensive overview of the \gls{mvtecad} dataset and the data split used in Experiment 3, described in Section~\ref{sec:exp3-fixednumimgs}. 
It displays the number of images for each anomaly type within every category. Additionally, it includes specific counts of anomalous images used for training purposes. 

As in Experiment 3, we establish a minimum number of anomalous images $\minnumimg$ in $\wvecgreedy$ and randomly select images from all anomaly types, this table shows the number of images allocated for the greedy eigencomponent selection. 
The remaining anomalous images not included in the greedy set $\wvecgreedy$ constitute the evaluation set $\wveceval$, and all the normal images from \gls{mvtecad}'s test set are shared by both.

\begin{table}[htbp]
    \singlespacing
    \small
    \setlength\tabcolsep{1pt}
  \centering
  \renewcommand*{\arraystretch}{.5}
  \begin{tabular*}{\textwidth}{
    @{\extracolsep{\fill}}
    p{0.1\linewidth}
    p{0.1\linewidth}
    p{0.1\linewidth}
    p{0.1\linewidth}
    p{0.1\linewidth}
    p{0.1\linewidth}
  }
    \hline
    \SetCell[c=1]{c} Category & \SetCell[c=1]{c} Test set split & \SetCell[c=1]{c} Images per anomaly type & \SetCell[c=1]{c} Greedy search split & \SetCell[c=1]{c} Evaluation split & \SetCell[c=1]{c} Train split (only normal images) \\
    \hline
    & & & & & \\
    & & & & & \\
\textbf{bottle}     & broken small           & 22                      & 5                   & 17               &                                     \\
                    & contamination          & 21                      & 5                   & 16               &                                     \\
                    & broken large           & 20                      & 5                   & 15               &                                     \\
                    & \textbf{Total}         & \textbf{63}             & \textbf{15}         & \textbf{48}      & \textbf{209}                        \\
                    & good                   & 20                      &                     &                  &                                     \\
                    &                        &                         &                     &                  &                                     \\
                    &                        &                         &                     &                  &                                     \\
\textbf{cable}      & missing\_wire          & 10                      & 2                   & 8                &                                     \\
                    & cable\_swap            & 12                      & 2                   & 10               &                                     \\
                    & bent\_wire             & 13                      & 2                   & 11               &                                     \\
                    & cut\_inner\_insulation & 14                      & 2                   & 12               &                                     \\
                    & poke\_insulation       & 10                      & 2                   & 8                &                                     \\
                    & missing\_cable         & 12                      & 2                   & 10               &                                     \\
                    & cut\_outer\_insulation & 10                      & 2                   & 8                &                                     \\
                    & combined               & 11                      & 2                   & 9                &                                     \\
                    & \textbf{Total}         & \textbf{92}             & \textbf{16}         & \textbf{76}      & \textbf{224}                        \\
                    & good                   & 58                      &                     &                  &                                     \\
                    &                        &                         &                     &                  &                                     \\
                    &                        &                         &                     &                  &                                     \\
\textbf{capsule}    & poke                   & 21                      & 3                   & 18               &                                     \\
                    & faulty\_imprint        & 22                      & 3                   & 19               &                                     \\
                    & squeeze                & 20                      & 3                   & 17               &                                     \\
                    & crack                  & 23                      & 3                   & 20               &                                     \\
                    & scratch                & 23                      & 3                   & 20               &                                     \\
                    & \textbf{Total}         & \textbf{109}            & \textbf{15}         & \textbf{94}      & \textbf{219}                        \\
                    & good                   & 23                      &                     &                  &                                     \\
                    &                        &                         &                     &                  &                                     \\
                    &                        &                         &                     &                  &                                     \\
\textbf{carpet}     & cut                    & 17                      & 3                   & 14               &                                     \\
                    & thread                 & 19                      & 3                   & 16               &                                     \\
                    & hole                   & 17                      & 3                   & 14               &                                     \\
                    & metal\_contamination   & 17                      & 3                   & 14               &                                     \\
                    & color                  & 19                      & 3                   & 16               &                                     \\
                    & \textbf{Total}         & \textbf{89}             & \textbf{15}         & \textbf{74}      & \textbf{280}                        \\
                    & good                   & 28                      &                     &                  &                                     \\
                    &                        &                         &                     &                  &                                     \\
                    &                        &                         &                     &                  &                                     \\
\textbf{grid}       & broken                 & 12                      & 3                   & 9                &                                     \\
                    & thread                 & 11                      & 3                   & 8                &                                     \\
                    & bent                   & 12                      & 3                   & 9                &                                     \\
                    & glue                   & 11                      & 3                   & 8                &                                     \\
                    & metal\_contamination   & 11                      & 3                   & 8                &                                     \\
                    & \textbf{Total}         & \textbf{57}             & \textbf{15}         & \textbf{42}      & \textbf{264}                        \\
                    & good                   & 21                      &                     &                  &                                     \\
                    &                        &                         &                     &                  &                                     \\
                    &                        &                         &                     &                  &                                     \\
\textbf{hazelnut}   & print                  & 17                      & 4                   & 13               &                                     \\
                    & hole                   & 18                      & 4                   & 14               &                                     \\
                    & cut                    & 17                      & 4                   & 13               &                                     \\
                    & crack                  & 18                      & 4                   & 14               &                                     \\
                    & \textbf{Total}         & \textbf{70}             & \textbf{16}         & \textbf{54}      & \textbf{391}                        \\
                    & good                   & 40                      &                     &                  &                                     \\
                    &                        &                         &                     &                  &                                     \\
                    &                        &                         &                     &                  &                                     \\
    
\textbf{leather}    & glue                   & 19                      & 3                   & 16               &                                     \\
                    & cut                    & 19                      & 3                   & 16               &                                     \\
                    & fold                   & 17                      & 3                   & 14               &                                     \\
                    & poke                   & 18                      & 3                   & 15               &                                     \\
                    & color                  & 19                      & 3                   & 16               &                                     \\
                    & \textbf{Total}         & \textbf{92}             & \textbf{15}         & \textbf{77}      & \textbf{245}                        \\
                    & good                   & 32                      &                     &                  &                                     \\
                    &                        &                         &                     &                  &                                     \\
                    &                        &                         &                     &                  &                                     \\
\textbf{metal\_nut} & color                  & 22                      & 4                   & 18               &                                     \\
                    & bent                   & 25                      & 4                   & 21               &                                     \\
                    & scratch                & 23                      & 4                   & 19               &                                     \\
                    & flip                   & 23                      & 4                   & 19               &                                     \\
                    & \textbf{Total}         & \textbf{93}             & \textbf{16}         & \textbf{77}      & \textbf{220}                        \\
                    & good                   & 22                      &                     &                  &                                     \\
                    &                        &                         &                     &                  &                                     \\
                    &                        &                         &                     &                  &                                     \\
    \end{tabular*}
    \phantomcaption
\end{table}

\begin{table}[htbp]
    \ContinuedFloat
    \singlespacing
    \small
    \setlength\tabcolsep{1pt}
  \renewcommand*{\arraystretch}{.5}
  \centering
  \begin{tabular*}{\textwidth}{
    @{\extracolsep{\fill}}
    p{0.1\linewidth}
    p{0.1\linewidth}
    p{0.1\linewidth}
    p{0.1\linewidth}
    p{0.1\linewidth}
    p{0.1\linewidth}
    @{}
  }
    \hline
    \SetCell[c=1]{c} Category & \SetCell[c=1]{c} Test set split & \SetCell[c=1]{c} Images per anomaly type & \SetCell[c=1]{c} Greedy search split & \SetCell[c=1]{c} Evaluation split & \SetCell[c=1]{c} Train split (only normal images) \\
    \hline
    & & & & & \\
    & & & & & \\

\textbf{pill}       & color                  & 25                      & 3                   & 22               &                                     \\
                    & scratch                & 24                      & 3                   & 21               &                                     \\
                    & contamination          & 21                      & 3                   & 18               &                                     \\
                    & combined               & 17                      & 3                   & 14               &                                     \\
                    & faulty\_imprint        & 19                      & 3                   & 16               &                                     \\
                    & pill\_type             & 9                       & 3                   & 6                &                                     \\
                    & crack                  & 26                      & 3                   & 23               &                                     \\
                    & \textbf{Total}         & \textbf{141}            & \textbf{21}         & \textbf{120}     & \textbf{267}                        \\
                    & good                   & 26                      &                     &                  &                                     \\
                    &                        &                         &                     &                  &                                     \\
                    &                        &                         &                     &                  &                                     \\
\textbf{screw}      & scratch\_head          & 24                      & 3                   & 21               &                                     \\
                    & thread\_top            & 23                      & 3                   & 20               &                                     \\
                    & scratch\_neck          & 25                      & 3                   & 22               &                                     \\
                    & thread\_side           & 23                      & 3                   & 20               &                                     \\
                    & manipulated\_front     & 24                      & 3                   & 21               &                                     \\
                    & \textbf{Total}         & \textbf{119}            & \textbf{15}         & \textbf{104}     & \textbf{320}                        \\
                    & good                   & 41                      &                     &                  &                                     \\
                    &                        &                         &                     &                  &                                     \\
                    &                        &                         &                     &                  &                                     \\
\textbf{tile}       & test,glue\_strip       & 18                      & 3                   & 15               &                                     \\
                    & test,gray\_stroke      & 16                      & 3                   & 13               &                                     \\
                    & test,oil               & 18                      & 3                   & 15               &                                     \\
                    & test,crack             & 17                      & 3                   & 14               &                                     \\
                    & test,rough             & 15                      & 3                   & 12               &                                     \\
                    & \textbf{Total}         & \textbf{84}             & \textbf{15}         & \textbf{69}      & \textbf{230}                        \\
                    & good                   & 33                      &                     &                  &                                     \\
                    &                        &                         &                     &                  &                                     \\
                    &                        &                         &                     &                  &                                     \\
\textbf{toothbrush} & defective              & 30                      & 15                  & 15               &                                     \\
                    & \textbf{Total}         & \textbf{30}             & \textbf{15}         & \textbf{15}      & \textbf{60}                         \\
                    & good                   & 12                      &                     &                  &                                     \\
                    &                        &                         &                     &                  &                                     \\
                    &                        &                         &                     &                  &                                     \\
\textbf{transistor} & cut\_lead              & 10                      & 4                   & 6                &                                     \\
                    & misplaced              & 10                      & 4                   & 6                &                                     \\
                    & damaged\_case          & 10                      & 4                   & 6                &                                     \\
                    & bent\_lead             & 10                      & 4                   & 6                &                                     \\
                    & \textbf{Total}         & \textbf{40}             & \textbf{16}         & \textbf{24}      & \textbf{213}                        \\
                    & good                   & 60                      &                     &                  &                                     \\
                    &                        &                         &                     &                  &                                     \\
                    &                        &                         &                     &                  &                                     \\
\textbf{wood}       & color                  & 8                       & 3                   & 5                &                                     \\
                    & liquid                 & 10                      & 3                   & 7                &                                     \\
                    & hole                   & 10                      & 3                   & 7                &                                     \\
                    & combined               & 11                      & 3                   & 8                &                                     \\
                    & scratch                & 21                      & 3                   & 18               &                                     \\
                    & \textbf{Total}         & \textbf{60}             & \textbf{15}         & \textbf{45}      & \textbf{247}                        \\
                    & good                   & 19                      &                     &                  &                                     \\
                    &                        &                         &                     &                  &                                     \\
                    &                        &                         &                     &                  &                                     \\
\textbf{zipper}     & combined               & 16                      & 3                   & 13               &                                     \\
                    & broken\_teeth          & 19                      & 3                   & 16               &                                     \\
                    & split\_teeth           & 18                      & 3                   & 15               &                                     \\
                    & squeezed\_teeth        & 16                      & 3                   & 13               &                                     \\
                    & rough                  & 17                      & 3                   & 14               &                                     \\
                    & fabric\_interior       & 16                      & 3                   & 13               &                                     \\
                    & fabric\_border         & 17                      & 3                   & 14               &                                     \\
                    & \textbf{Total}         & \textbf{119}            & \textbf{21}         & \textbf{98}      & \textbf{240}                        \\
                    & good                   & 32                      &                     &                  &                                     
  \end{tabular*}
    \caption{MVTec-AD Image Count Details (for Experiment 3)}
    \label{tab:num_images}
\end{table}


\end{document}